%% file: main.tex
\documentclass{isprs} 
\usepackage{subfigure}
\usepackage{graphicx}
\usepackage[table,xcdraw]{xcolor}
\usepackage{tabularx}
\usepackage{amsmath}
\usepackage{amssymb}
\usepackage{booktabs}
\usepackage{multirow}
\usepackage{setspace}
\usepackage{geometry} 
\usepackage{pifont}
\usepackage{xcolor}
\usepackage{url}

\usepackage[utf8]{inputenc}
\usepackage[labelsep=period]{caption}  
\usepackage[british]{babel} 
\usepackage[hang]{footmisc}
\usepackage{arydshln} 
\usepackage{todonotes}
\usepackage[switch]{lineno} 
\usepackage{etoolbox}      
\usepackage{xcolor}
\usepackage[normalem]{ulem}

\makeatletter

\patchcmd\makeLineNumber
  {\rlap}
  {\if@twocolumn\ifodd\c@page\hskip\columnwidth\fi\fi\rlap}
  {}{}
\makeatother

\begin{document}
\title{TimeSenCLIP: A Time Series Vision-Language Model for Remote Sensing}

\date{}

\author{Pallavi Jain\textsuperscript{1, 2, 6}, Diego Marcos\textsuperscript{2, 6}, Dino Ienco\textsuperscript{2, 3, 5, 6}, Roberto Interdonato\textsuperscript{2, 4, 5, 6}, and Tristan Berchoux\textsuperscript{1,5,6}}

\address{
\textsuperscript{1}Mediterranean Agronomic Institute of Montpellier - CIHEAM-IAMM,
\textsuperscript{2} Inria,  \textsuperscript{3} INRAE, \\ \textsuperscript{4} Cirad, \textsuperscript{5} UMR TETIS, \textsuperscript{6} Univ. of Montpellier, Montpellier, France \\
{\tt\small \{pallavi.jain, diego.marcos, dino.ienco, roberto.interdonato\}@inria.fr }\\
{\tt\small  berchoux@iamm.fr}\\
}

\abstract{
Vision-language models (VLMs) have shown significant promise in remote sensing applications, particularly for land-use and land-cover (LULC) mapping via zero-shot classification and retrieval. However, current approaches face several key challenges, such as the dependence on caption-based supervision, which is often not available or very limited in terms of the covered semantics, and the fact of being adapted from generic VLM architectures that are suitable for very high resolution images. Consequently, these models tend to prioritize spatial context over spectral and temporal information, limiting their effectiveness for medium-resolution remote sensing imagery.  \\

In this work, we present TimeSenCLIP, a lightweight VLM for remote sensing time series, using a cross-view temporal contrastive framework to align multispectral Sentinel-2 time series with geo-tagged ground-level imagery, without requiring textual annotations. Unlike prior VLMs, TimeSenCLIP emphasizes temporal and spectral signals over spatial context, investigating whether single-pixel time series contain sufficient information for solving a variety of tasks. \\

Our approach is trained on the LUCAS and Sen4Map datasets and evaluated across four main mapping tasks: land cover, land use, habitat mapping and crop type classification. 
The CLIP text encoder can be used to probe the learned representations using semantically meaningful categories, enabling effective zero-shot generalization without task-specific text supervision.
We further extend our evaluation to bioregions mapping and country-level image retrieval. Although coarse, these tasks are valuable for probing whether the model captures geographically meaningful representations, such as regional climate regimes, vegetation patterns, and land-use structures. TimeSenCLIP achieves consistently better performance than existing CLIP-based remote sensing models in both zero-shot classification and cross-modal retrieval. Notably, single-pixel multispectral time series variants remain highly competitive, particularly with extended temporal coverage, demonstrating that temporal–spectral dynamics can compensate to a substantial degree for the reduced spatial footprint. \\

While larger spatial patches still offer advantages for tasks where spatial patterns are inherently informative, such as ecosystem type classification, the results suggest that single-pixel multispectral time series can provide effective remote sensing vision–language pipelines, supporting scalable and efficient modelling in scenarios where large spatial tiles or extensive textual annotations are impractical. Code is available at \url{https://github.com/pallavijain-pj/TimeSenCLIP}.
}

\keywords{VLMs, Time Series, Multispectral, Remote Sensing, Sentinel-2, Cross-View, Contrastive learning }

\maketitle

\section{Introduction}\label{sec:intro}
\sloppy

 Remote sensing technology is central to large-scale environmental monitoring, providing critical insights into land cover change, ecosystem health, biodiversity assessment, and agricultural productivity~\cite{li2014remote,soubry2021systematic}. The use of machine learning for the enhanced analysis of  satellite data has enabled increasingly automated and accurate classification systems. However, the challenge of scaling out across diverse ecosystems, sensing modalities, and geographical regions remains largely unresolved, particularly when computational efficiency and generalization to unseen classes are required~\cite{reichstein2019deep,zhu2017deep,qiu2024few,tan2024review}.

Recent advances in vision-language models (VLMs) have opened new possibilities in this domain. Models like CLIP \cite{radford2021learning}, which learn aligned embeddings between images and natural language, have shown remarkable zero-shot generalization across a wide range of vision tasks such as object recognition, semantic segmentation, image captioning, and cross-modal retrieval~\cite{tschannen2025siglip}. When adapted to remote sensing, VLMs allows for open-vocabulary classification and text-driven retrieval of overhead imagery data using natural language queries (e.g., ``arid farmland with sparse vegetation" or ``oil palm plantations near waterways"), significantly reducing the need for exhaustive, class-specific supervision \cite{dhakal2023sat2cap,jain2025senclip,mall2023remote,vivanco2023geoclip,klemmer2025satclip}. 

\begin{figure}
        \centering
        \includegraphics[width=\linewidth, scale=3]{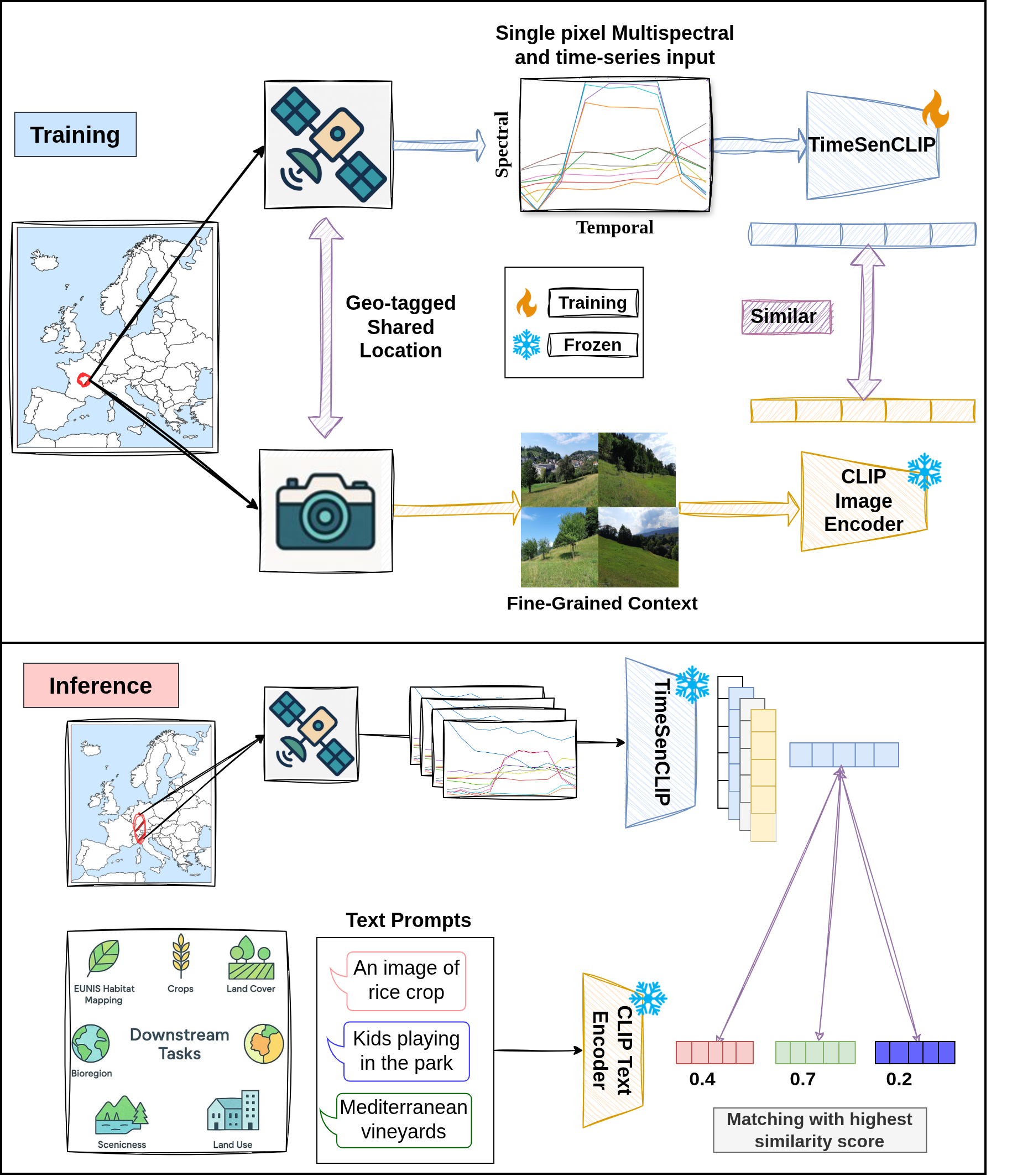}
        \caption{TimeSenCLIP Pipeline Illustration: Satellite Sentinel-2 single pixel multispectral time series are aligned with geo-tagged ground-level images through cross-view learning, enabling the model to capture fine-grained ecological semantics without relying on large spatial context or text supervision. }
        \label{fig:illustration_pipeline}
    \end{figure}
This capability is particularly appealing for ecosystem mapping, which extends beyond traditional land-use and land-cover (LULC) classification (e.g., ``urban" or ``cropland"),
by combining diverse biophysical information to produce thematic maps that depict the spatial distribution, scale, and ecological linkages of natural systems within a given landscape (e.g., for mapping biodiversity, habitat, and ecological regions) \cite{keith2022function,lausch2016linking}. Such distinctions are often fine-grained, context-dependent, and not well captured by fixed-class LULC taxonomies. Addressing this complexity typically requires either labor-intensive expert annotation or flexible models capable of adapting to unseen descriptions. VLMs bridge this gap by interpreting open-ended text prompts to identify nuanced classes without retraining (eg.``Intensively managed broadleaf forest dominated by Fagus sylvatica" versus simply ``forest") ~\cite{jain2025senclip}.

Despite these advantages, extending VLMs to remote sensing data faces several key challenges. 
\textbf{First}, many existing remote-sensing VLMs depend heavily on text-labeled training data or curated caption datasets. Such annotations are costly, exhibit vocabulary bias, and remain insufficient for fine ecosystem granularity, regional habitat terminology, or visually subtle classes. As a result, current text-supervised models often fail to generalize beyond the concepts in their rigid text training set. 
\textbf{Second}, most remote sensing VLMs are adapted from generic VLMs and adopt large spatial input images (e.g., 200×200 to 512×512 px)~\cite{zhang2016deep,zhu2017deep}, rendering them more suitable for for very high resolution imagery than medium resolution like Sentinel-2. While spatial context is often valuable, a high pixel count with medium resolution imagery results in a very large context that can be unreliable in fragmented or heterogeneous landscapes, particularly in ecological settings, where cloud contamination, mixed pixels, seasonal transitions~\cite{koldasbayeva2024challenges} or weak spatial autocorrelation reduce the discriminative power of spatial context. 
\textbf{Third}, a large body of remote sensing research shows that, at the medium spatial resolution such as Sentinel-2, many ecological, agricultural, and phenological classes are more distinctly characterized by multispectral and temporal signatures than by spatial patterns~\cite{zhong2019deep,russwurm2018multi,lausch2016linking,pettorelli2016framing,pesaresi2022functional}. Yet, most VLM adaptations rely primarily on static RGB representations, leaving spectral and temporal information underexploited.
\textbf{Finally}, the aquisition, computational and storage costs associated with high-resolution patches, multimodal fusion, and text-based supervision limit the practicality of VLMs for large-scale or long-term monitoring.

To address these limitations, we propose TimeSenCLIP, a lightweight model, that adapts the VLM paradigm to operate without unnecessary spatial context and without the need for additional text annotations. We specifically investigate whether the spectral-temporal signature of a single Sentinel-2 pixel contains sufficient semantic information to support accurate land-use, land-cover, and ecosystem classification. This choice directly challenges the assumption that large spatial neighborhoods are necessary, emphasizing the discriminative potential of spectral-temporal dynamics.

Rather than training with text labels or captions, TimeSenCLIP adopts a cross-view learning paradigm as shown in Figure~\ref{fig:illustration_pipeline} (top), that aligns satellite image time series with geo-tagged ground-level photographs. 
This type approach can leverage any large collection of geotagged photos, such user generated ones on Flickr, used by Sat2Cap~\cite{dhakal2023sat2cap} and GRAFT~\cite{mall2023remote}, or the large scale LUCAS survey used by SenCLIP~\cite{jain2025senclip}.
This setup helps reduce spurious correlations that can arise from caption-based training, due to the limitations of building a caption dataset, and encourages the model to capture more fine-grained ecological signals, such as vegetation structure, land use intensity, or habitat composition. 
At inference time, satellite time series embeddings produced by TimeSenCLIP can be compared with textual descriptions of the classes of interest, as seen in Figure~\ref{fig:illustration_pipeline} (bottom).
By leveraging the rich language understanding of CLIP through ground-level images, TimeSenCLIP can go beyond generic text prompts based on class names or conventional aerial- or satellite-view textual prompts (e.g., \textit{"a centered satellite image of a residential building"} or \textit{"an aerial view of a plantation"}), enabling generalization to nuanced descriptions of land-use and ecosystem type from a ground-level perspective.

Unlike prior work, TimeSenCLIP explicitly incorporates Sentinel-2’s multispectral and temporal dimensions, demonstrating their usefulness for tasks such as crop mapping, bioregion delineation, and habitat mapping. Because temporal-spectral data already provides the necessary signal to solve these tasks, we use a single-pixel time series as input to keep the model lightweight and computationally efficient.
This potentially entails a mismatch between the objects depicted by each modality, with ground-level images capturing details of objects in the scene and a large spatial context, albeit in a single temporal moment, while Sentinel-2 time series capture the dynamics of the average reflectance of a $100 m^2$ surface.
However, our results suggest that both these different views of a location can be aligned meaningfully and are able to capture the essential land-use characteristics.

Our main contributions are: 

\begin{enumerate}
    \item We introduce TimeSenCLIP, a cross-view learning approach aligning satellite spectral-temporal observations with ground-level semantics using ground-level photos and a text-aligned vision model, eliminating dependence on caption-based supervision.
    \item We investigate the role of temporal modelling in text-aligned satellite representations, disentangling the contributions of temporal, spectral, and spatial information and showing that temporal–spectral dynamics are critical for fine-grained ecological and land-use understanding.
    \item We provide a comprehensive assessment of our framework across diverse tasks, including LULC classification, habitat mapping, bioregion mapping, crop type identification, and scenicness estimation, showing consistent zero-shot performance while maintaining computational efficiency through single-pixel temporal modelling.
\end{enumerate}

Together, these contributions highlight the practical potential of temporally and spectrally informed vision–language models for scalable, fine-grained zero-shot mapping in remote sensing.

\section{Related Works}\label{sec:related}

Our approach builds upon recent advances in vision-language alignment, remote sensing foundation models, and multivariate time series learning , but departs from prior work by leveraging satellite time series and ground level imagery alignment, without requiring paired-text supervision.

\subsection{VLMs in Remote Sensing}
VLMs such as CLIP~\cite{radford2021learning} have significantly advanced zero-shot performance by projecting images and text into a shared embedding space, enabling flexible and scalable classification across diverse domains. Inspired by these advances, recent efforts have adapted VLMs to the remote sensing domain. Notable examples include GeoRSCLIP~\cite{zhang2023rs5m}, SkyCLIP~\cite{wang2024skyscript}, and RemoteCLIP~\cite{liu2023remoteclip}, which leverage satellite imagery and natural language to perform open-vocabulary land cover classification. These models demonstrate strong generalization by pairing large-scale remote sensing imagery with curated textual descriptions. Beyond classification, models like GeoPixel~\cite{shabbir2025geopixel} extend vision-language alignment to the pixel and patch level, enabling fine-grained captioning and grounding over high-resolution Earth observation data. 
However, such models often assume that semantic understanding can only emerge from larger spatial contexts. This is reflected in the use of large image patches (e.g., $224 \times 224$ pixels), which implicitly rely on local spatial continuity for semantic inference. While effective in many structured environments, this assumption becomes problematic in ecologically diverse or fragmented landscapes, such as mixed-crop, hedgerow mosaics, agroforestry systems, seasonal floodplain wetlands, or shrub-steppe regions, where spatial coherence is weak, discontinuous, or even misleading. A second limitation lies in their reliance on paired textual description supervision; training typically depends on manually curated image–text datasets (e.g., RS5M~\cite{zhang2023rs5m}), which are costly to scale and often suffer from vocabulary bias, especially in underrepresented ecological contexts.

Our work challenges these assumptions by exploring whether rich semantics can emerge from single pixel represented solely by their spectral-temporal signatures. We demonstrate that, without relying on large spatial context or curated text prompts, small inputs encode sufficient information to support ecosystem retrieval and alignment.

\subsection{Cross-View Ground–Satellite Alignment}
Cross-view supervision has emerged as a compelling strategy for learning geospatial representations by aligning satellite information and ground-level imagery. Models such as Sat2Cap~\cite{dhakal2023sat2cap}, SenCLIP~\cite{jain2025senclip}, and GRAFT~\cite{mall2023remote} leverage contrastive or generative objectives to bridge the domain gap between views. SenCLIP, in particular, introduces an attention pooling mechanism across directional ground-level images, processed via a frozen CLIP encoder and aligns them with Sentinel-2 inputs, inspiring the ground encoder design in our model. Similarly, GAIR~\cite{liu2025gair} applies hierarchical fusion of satellite and ground imagery to learn coherent geospatial embeddings for downstream tasks.

In contrast to these approaches, we propose to align spectral-temporal pixel from Sentinel-2 with ground-level features through direct contrastive learning, avoiding any reliance on spatial priors, or language-based supervision. This enables learning from extremely small spatial inputs (as little as single pixel), while retaining semantic richness from high-resolution ground imagery. Our model thereby offers a
supervision efficient alternative for geospatial representation learning, applicable across ecological, agricultural, and other land-use domains.

\subsection{Multispectral and Temporal Remote Sensing Models}
Temporal modeling is a cornerstone of remote sensing research, particularly for applications such as ecological and agricultural applications.
Convolution-based architectures like TempCNN \cite{pelletier2019temporal} and hybrid CNN–attention models such as L-TAE \cite{garnot2020ltae} and ConvTran \cite{ConvTran2023} have demonstrated strong performance on time series satellite data. However, their effectiveness often depends on extensive preprocessing and regularly sampled time series.

Recent advances in spectral-temporal VLMs for remote sensing have further explored text alignment using such models. Notable examples include Llama3-MS-CLIP, which extends RGB inputs to multispectral patches \cite{marimo2025beyond}; GeoLLAVA, which treats time series as video–language pairs \cite{elgendy2024geollava}; and EarthDial, which employs a two-stage RGB-to-multispectral and temporal fine-tuning scheme \cite{soni2024earthdial}. While these models represent important progress towards truly spatio-spectral VLMs for Earth observation, they remain limited by their reliance on RGB priors, fixed temporal windows, or the need for extensive supervision. In contrast, our transformer-based framework directly processes raw spectral-temporal cubes, jointly modeling multispectral (10 Sentinel-2 bands) and multi-temporal information spatial upsampling, or multi-stage pretraining.

\subsection{Ecological Representation and Remote Sensing Foundations}

Recent work has advanced ecological representation learning by aligning remote sensing imagery with external ecological knowledge, including it in the form of text. For example, EcoWiKiRS~\cite{zermatten2025ecowikirs} and WildSAT~\cite{daroya2024wildsat} leverage species occurrence data and habitat preferences, while TaxaBind~\cite{sastry2025taxabind} aligns a variety of ecologically relevant modalities, including satellite imagery. Other efforts focus on multi-sensor remote sensing data fusion for ecosystem modelling~\cite{wang2025towards}.

Our approach contributes an alternative, without requiring class labels, taxonomies, or curated prompts, by aligning Sentinel-2 spectral-temporal pixel with image-level supervision from ground-level imagery. It avoids spatial priors or dense annotations (e.g., maps), and instead learns directly from raw satellite image time series pixels through cross-view supervision.

This setup enables semantic representations to emerge from spectral-temporal structure alone, supporting a broader shift toward scalable, data-driven ecosystem modelling in remote sensing VLMs.

\section{Method}\label{sec:method}

\begin{figure*}[htbp]
        \centering
        \includegraphics[ scale=0.55]{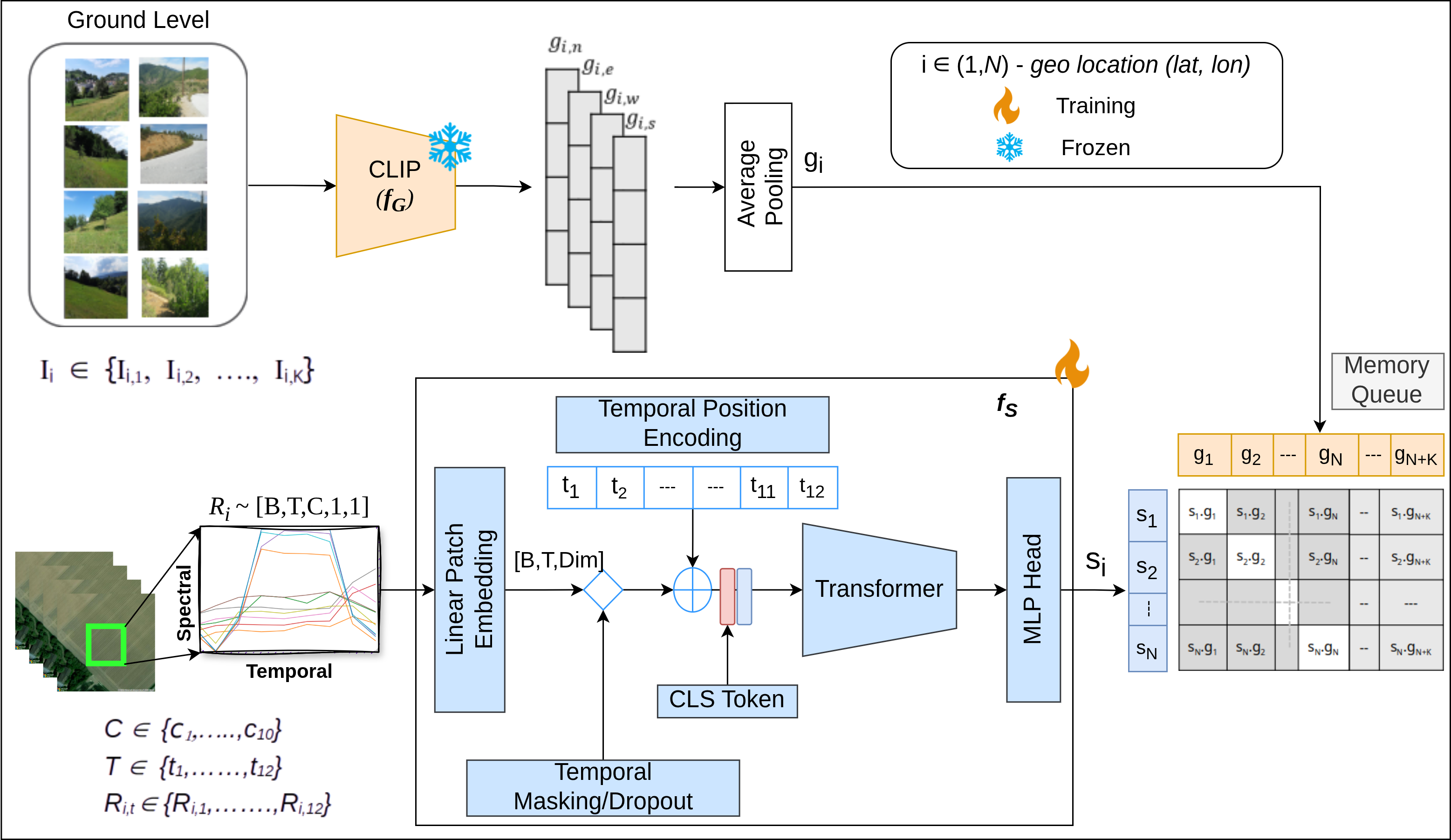}
        \caption{TimeSenCLIP Model Training: Spectral-temporal patches from Sentinel-2 are aligned with ground-level CLIP features using contrastive learning. The satellite encoder learns from minimal spatial input via a transformer, while a memory queue enables efficient negative sampling.}
        \label{fig:model_arch}
    \end{figure*}
Our approach, TimeSenCLIP (Temporal VLM for Sentinel-2 imagery), illustrated in Figure~\ref{fig:model_arch}, builds on the prior satellite–ground alignment works of SenCLIP~\cite{jain2025senclip}. 
SenCLIP leverages CLIP-based encoders for both ground-level and satellite modalities, making it suitable only for RGB imagery.

With TimeSenCLIP, we use a frozen CLIP image encoder only for the ground-level modality and train a multivariate time series model from scratch for the satellite modality, bridging spectral-temporal satellite observations with semantically rich ground-level representations.

\paragraph{Overview} During training, our approach makes use of two main modules:
\begin{itemize}
    \item Ground-level encoder: The frozen CLIP image encoder provides semantically rich embeddings that act as a text-aligned proxy target. Multiple directional images are aggregated via attention pooling to focus on the most informative perspectives. This module is only used for training. 
    \item Satellite encoder: A transformer-based architecture captures long-range spectral-temporal dependencies, accommodating monthly, quarterly, or annual time series. Via contrastive training, this model learns to provide a text-aligned representation of the time series. The resulting model is the one used at inference in order to perform zero-shot classification of Sentinel-2 time series.

\end{itemize}

Our design is guided by two main objectives: 
(1) to leverage the CLIP encoder as a semantic anchor, transferring knowledge from natural image-language domains into remote sensing; and
(2) to capture spectral-temporal dependencies in Sentinel-2 data using a transformer-based encoder capable of handling monthly, quarterly and annual time series.

\subsection{CLIP Overview}
CLIP (Contrastive Language–Image Pretraining)~\cite{radford2021learning} is a large-scale vision-language model trained on hundreds of millions of image-text pairs. It learns to align images and their corresponding textual descriptions in a shared embedding space using a contrastive objective, such that matching image-text pairs have high cosine similarity and non-matching pairs have low similarity. 

Formally, for sample $i$, let $\mathbf{v}_i$ and $\mathbf{t}_i$ denote the L2-normalized image and text embeddings from image encoder $f_\text{img}$ and text encoder $f_\text{text}$:

\begin{equation}
\mathbf{v}_i = \frac{f_\text{img}(I_i)}{\|f_\text{img}(I_i)\|_2}, \quad 
\mathbf{t}_i = \frac{f_\text{text}(T_i)}{\|f_\text{text}(T_i)\|_2}.
\end{equation}

The cosine similarity between embeddings is then the dot product $\mathbf{v}_i \cdot \mathbf{t}_j$.

Maximizing similarity for matched pairs and minimizing it for
non-matched pairs produces a semantically and visually rich latent space that generalizes across natural and human-made scenes.

This is done using the InfoNCE loss, which is formulated as:

\begin{multline}
\mathcal{L}_{\text{InfoNCE}}(\mathbf{x},\mathbf{x}^+,\{\mathbf{x}_j^-\}) = \\ -\log \frac{\exp(\mathbf{x} \cdot \mathbf{x}^+ / \tau)}{\exp(\mathbf{x} \cdot \mathbf{x}^+ / \tau) + \sum_{j=1}^J \exp(\mathbf{x} \cdot \mathbf{x}^-_j / \tau)},
\end{multline}
where $\mathbf{x}$ is the anchor sample, $\mathbf{x}^+$ a positive sample, possibly from a different modality, and $\{\mathbf{x}_j\}$ represents a set of negative samples, and $\tau$ is a learnable temperature controlling the concentration of the similarity distribution. In CLIP, where both an image and a text models are learned jointly, the total loss would be $\mathcal{L}_\text{CLIP} = \sum_i \mathcal{L}_{\text{InfoNCE}}(\mathbf{v}_i,\mathbf{t}^+_i,\{\mathbf{t}_j^-\}_i) + \mathcal{L}_{\text{InfoNCE}}(\mathbf{t}_i,\mathbf{v}^+_i,\{\mathbf{v}_j^-\}_i) $

\subsection{Model Architecture}
During training, TimeSenCLIP makes use of an image encoder and a time-series encoder: a frozen CLIP image encoder $f_G$ for the ground-level images and a trainable, transformer-based, satellite time-series encoder $f_S$.
After training, only $f_S$ is retained, and it can be used together with the CLIP text encoder $f_T$ in order to infer the alignment between a satellite time series and any textual prompt.

\paragraph{Ground-level Encoder (CLIP).} 

For each geographic location, multiple ground-level images can be available: 
$\{I_{1}, \dots, I_{K}\}$. These $K$ images are individually processed by the pretrained CLIP image encoder to produce the corresponding feature embeddings 
$\{\mathbf{g}_{1}, \dots, \mathbf{g}_{K}\}$. These embeddings are then aggregated via an average pooling to yield a single ground-level descriptor.

\paragraph{Satellite Encoder.}
For the \textbf{Sentinel-2 time series}, we design a transformer-based model $f_S(\cdot)$ that takes as input a spectral-temporal cube $\mathbf{R} \in \mathbb{R}^{T \times C \times H \times W}$, where $T$ denotes the number of temporal frames, $C$ is the number of spectral bands, and $H \times W$ is the spatial extent. In the single-pixel setting, 
($H = 1$, $W = 1$), whereas larger spatial dimensions can be used to incorporate additional spatial context. Each time slice $\mathbf{R}_t \in \mathbb{R}^{C \times H \times W}$ is flattened and projected with a linear layer:

\begin{equation}
\mathbf{r}_t = \text{Linear}\left(\text{Flatten}(\mathbf{R}_t)\right), \quad t = 1, \dots, T
\end{equation}

To improve robustness to missing or irregularly sampled temporal data, we apply temporal augmentations during training, including random quarterly masking and median pooling of time steps, simulating partially observed sequences (discussed in Section~\ref{sec:temp_aug}).

Learnable temporal position embeddings $\mathbf{p}_t$ are added to each time step, and a learnable \textit{class token} $\mathbf{p}_\text{cls}$ is prepended to summarize the entire spectral-temporal trajectory:
\begin{equation}
\mathbf{Z} = \left[\mathbf{p}_\text{cls}; \mathbf{r}_1 + \mathbf{p}_1; \dots; \mathbf{r}_T + \mathbf{p}_T\right]
\end{equation}

The tokenised sequence is then processed by the transformer-based encoder:
\begin{equation}
\mathbf{\hat{s}} = f_S(\mathbf{Z}).
\end{equation}
The output corresponding to the class token $\mathbf{p}_\text{cls}$ is extracted as the initial satellite representation and further processed by an MLP.

The choice of a vanilla transformer backbone is motivated by its ability to model long-range dependencies through global self-attention, which is particularly beneficial for spectral-temporal data. Unlike CNN-based architectures that rely on fixed local receptive fields, the transformer captures dynamic inter-band relationships and cross-seasonal temporal patterns within a unified representation space, improving flexibility across different sensors and temporal resolutions.

\paragraph{Cross-modal Alignment with Contrastive Learning.} 
To align satellite spectral-temporal embeddings with ground-level visual embeddings, we adopt an InfoNCE-based \cite{oord2018representation} contrastive learning framework. Let $\mathbf{s}$ denote a satellite embedding and $\mathbf{g}$ its corresponding ground-level embedding. Both embeddings are L2-normalized:

\begin{equation}
\mathbf{s} = \frac{\mathbf{\hat{s}}}{\|\mathbf{\hat{s}}\|_2}, \quad \mathbf{g} = \frac{\mathbf{\hat{g}}}{\|\mathbf{\hat{g}}\|_2}.
\end{equation}

The positive pair $(\mathbf{z}_S, \mathbf{z}_G)$ represents the satellite and ground images from the same location, while negative samples are drawn from a large MoCo-style~\cite{he2020momentum} memory bank $\mathcal{Q}$ that stores previously seen ground embeddings. This design allows the model to scale efficiently without relying on extremely large batch sizes. The final loss we use is thus:

\begin{multline}
    \mathcal{L}_\text{TimeSenCLIP}(\{\mathbf{s}_i\},\{\mathbf{g}_i\} , \mathcal{Q}) = \\ \sum_i \mathcal{L}_{\text{InfoNCE}}(\mathbf{s}_i,\mathbf{g}^+_i,\{\mathbf{g}_j^-\}_i \cup \mathcal{Q}).
\end{multline}
 Here, $\mathbf{g}^+_i$ denotes the ground image corresponding to the same location $i$ as the satellite time series $\mathbf{s}_i$, whereas $\mathbf{g}_j^-$ represents a negative samples, any other locations within the batch and $\mathcal{Q}$ is a memory bank storing ground embeddings from previously seen locations.
 
 Minimizing this loss encourages the satellite embedding to be close to its corresponding ground embedding while pushing it away from negatives, effectively aligning cross-modal representations in the CLIP latent space. This approach provides both robust alignment and computational scalability for large spectral-temporal datasets.

\subsection{Temporal Augmentation Strategy} 
\label{sec:temp_aug}
To account for missing or irregular observations in satellite time series due to cloud cover, seasonal occlusions, or sensor issues, we apply stochastic temporal masking during training. These augmentations are applied on-the-fly to the linear patch embeddings with a 50\% probability per batch, ensuring diverse perturbations across epochs. 

We randomly applied one of several temporal augmentations with equal probability, simulating partial or aggregated temporal observations:
        \begin{itemize}
        \item Median Pooling: Collapses all temporal frames into a single median vector, simulating temporally aggregated data (e.g., annual composites). This encourages robustness to varying temporal granularity while preserving spectral-temporal patterns.
        \item Random Quarter Mask: Randomly masks a contiguous subset of temporal frames (e.g., quarterly) to simulate realistic seasonal gaps caused by cloud cover or missing acquisitions.
        \item Random Temporal Mask: Randomly masks between 1 and 11 temporal frames, ensuring at least one frame is retained. This teaches the model to handle variable-length sequences and temporal inconsistencies.
        
    \end{itemize}

These three \textbf{training-only strategies} allow the model to learn robust spectral-temporal representations under incomplete or aggregated inputs. At inference, the model can flexibly operate on \textbf{monthly, quarterly, or single-aggregated sequences} (discussed in Section~\ref{sec:temp_agg}), enabling evaluation under different temporal resolutions without retraining.

\input{table_prompt_example}
\subsection{Prompt Design} \label{subsec:prompts}
For zero-shot evaluation, we construct three types of textual prompts: (i) \textbf{class names}, (ii) \textbf{generic} templated prompts of the form “a centered satellite image of \{class name\}”, and (iii) \textbf{class descriptive} natural language prompts. Five descriptive variations per class were generated using GPT-4, following the style of SenCLIP~\cite{jain2025senclip}.

Using multiple prompt types allows us to evaluate the model’s ability to capture fine-grained semantics: class names provide minimal, direct information; template prompts add context to help differentiate visually similar categories; and class descriptive prompts offer alternative phrasings that account for natural variation in how a class may be described. Because TimeSenCLIP is trained with ground-level images, it captures broader contextual information about locations. This enables the model to better leverage descriptive prompts, often yielding higher zero-shot performance than using only class names or templates. Moreover, as ground-level image representations rather than textual captions are used during training, the model is not limited to the usual remote sensing terminology and can generalize to natural language descriptions better adapted to characterize outdoor scenes.

We performed an exploration study to assess the contribution of each prompt type independently, reporting performance using only class names, only template prompts, or only descriptive prompts. For descriptive prompts, late prompt ensembling is applied on the CLIP embeddings by averaging the embeddings of the five prompts per class~\cite{menon2022visual,roth2023waffling} and then computing the similarity with the satellite embeddings, producing a single class prediction for each satellite image time series. Examples of descriptive prompts are provided in Table~\ref{tab:prompt_example}.

\section{Dataset}
\subsection{Cross-View Training Dataset}

\paragraph{Ground-Level Images.} The LUCAS (Land Use/Cover Area frame Survey) is a European Union-wide in-situ data collection campaign designed to systematically monitor land use and land cover (LULC) across Europe. The survey, conducted in 2018 \cite{d2020harmonised}, encompasses approximately 235,000 georeferenced sampling points distributed across 28 (EU member states and United Kingdom) countries. Each site is documented with high-resolution ground-level images taken in four cardinal directions (north, east, south, and west), accompanied by detailed annotations of land use, land cover (LULC), and crop types. 

\paragraph{Sentinel-2 Time Series Data.} 
\input{table_data_stats}

We employ the Sen4Map dataset~\cite{sharma2024sen4map}, which provides co-registered multi-spectral and multi-temporal Sentinel-2 observations aligned with the LUCAS (2018) in-situ survey points across Europe. 
Each sample represents a 64$\times$64 pixel Sentinel-2 patch centered on a LUCAS location and includes 10 spectral bands at 10\,m and 20\,m spatial resolutions, aggregated into monthly median composites to ensure cloud-free annual temporal coverage, resulting in 12 time steps per year. 
We used the Sen4Map random split, comprising 140k training, 30k validation, and 50k test samples.  We utilised train split for contrastive pre-training of  the model and evaluated our model on test split. 

Although the dataset provides 64$\times$64 spatial patches, TimeSenCLIP operates primarily on single-pixel ($1\times1$) inputs extracted from the centre of each patch. This design choice minimizes computational overhead while ensuring that each location is represented solely by its temporal–spectral signature, rather than by larger spatial context. For ablation studies, we also experiment with $5\times5$ and $9\times9$ patches to assess the influence of spatial context.

\subsection{Evaluation Tasks and Datasets} 

We evaluate TimeSenCLIP on multiple downstream tasks derived from the same geo-referenced dataset, encompassing land classification, habitat mapping, and perceptual quality prediction.

\paragraph{Land Use / Land Cover and Crop Type.}

We use coarse-grained land use and land cover (LULC) labels derived from the LUCAS field survey\footnote{\url{https://ec.europa.eu/eurostat/documents/205002/8072634/LUCAS2018-C3-Classification.pdf\#page=10.09}}, which follows the CORINE classification system. These labels include land cover categories such as artificial surfaces, croplands, grasslands, wetlands and forest, and define a multi-class classification task, based on in-situ observations across Europe. The Level 0 taxonomy comprises 8 land cover and 16 land use classes, while Level 1 further distinguishes croplands into 38 specific crop types. Figure \ref{fig:test_data_dist} in appendix showcases the Level-0 land cover class distribution across the EU within evaluation data split.

\paragraph{Biogeographical Zones.}
The biogeographical regions of Europe were obtained from the European Environmental Agency (EEA)\footnote{\url{https://www.eea.europa.eu/en/analysis/maps-and-charts/biogeographical-regions-in-europe-2}}~\cite{eea_bioregions_2016}.
This dataset divides Europe into 11 ecologically distinct zones (e.g., Alpine, Boreal, Mediterranean) based on climatic, topographic, and vegetational characteristics. Each LUCAS site was spatially assigned to its corresponding biogeographical zone.
Among the 11 regions, Arctic, Anatolian, Macaronesian, and Outside Europe were excluded, as these zones were not represented in the LUCAS geotag coverage.

\paragraph{Habitat Mapping.}
The European Nature Information System (EUNIS) habitat classification, provided by the EEA~\cite{eea_ecosystem_types_2019}, offers a continent-wide ecosystem map at 100 m spatial resolution\footnote{\url{https://sdi.eea.europa.eu/catalogue/srv/api/records/7c0cf3f2-ab54-4cd0-a635-b322df7197f6}}.
We employ Level 2 of the EUNIS hierarchy, comprising 44 terrestrial habitat types (e.g., Mixed deciduous and coniferous woodland, Arable land and market gardens). These fine-grained ecological categories enrich the dataset by capturing detailed habitat characteristics that complement broader land-use labels.

\paragraph{Scenicness Prediction.}

Scenicness prediction evaluates the ability of TimeSenCLIP to transfer learned representations to subjective, human-centered perceptual tasks, extending the typical classification-focused applications of remote sensing vision–language models (VLMs) to regression tasks. This task is formulated as a visual regression problem aimed at predicting the perceived beauty of landscapes~\cite{workman2017understanding,levering2021relation,levering2024prompt}.
We use the ScenicOrNot (SoN) dataset~\cite{seresinhe2015quantifying}, which contains geotagged ground-level images across the United Kingdom (UK) rated by human annotators on a scale from 1 to 10. These crowd-sourced scores serve as ground-truth labels for perceived aesthetic quality. Predicting scenicness is challenging due to subjective variability, sparse spatial coverage, and the temporal–spectral complexity of satellite observations.
To address this, we adopt the scenicness-oriented textual prompts from ~\cite{levering2024prompt} (e.g., ``a busy highway'' vs.\ ``a mountain'') and apply late prompt ensembling to capture concept-level aesthetics.
For evaluation, we use 2,411 Sen4Map samples from the UK, aligned to the nearest SoN image (within 100~m), maintaining the temporal and spectral information of the satellite time series. This setup demonstrates that TimeSenCLIP can generalize from semantic alignment to regression-based perceptual tasks, broadening the scope of remote sensing VLM applications.

Together, these datasets enable comprehensive evaluation of the model’s capacity for landscape characterization, capturing both LULC patterns and broader ecological attributes across diverse European regions. Table~\ref{tab:data_stats} provides an overview of the evaluation dataset, detailing the number of classes per label type and highlighting the eight most frequent classes with their corresponding proportions.

\section{Experimental Setup}
\subsection{Implementation Details}
\label{sec:imp_details}

\paragraph{Ground-Level Encoder.}
We use the ViT-B/32 CLIP model~\cite{radford2021learning}, pretrained on natural images, as the ground-level encoder. Each LUCAS ground-level image is independently passed through the frozen CLIP encoder, producing a 512-dimensional embedding.

\paragraph{Satellite Time Series Encoder.}
The satellite branch processes Sentinel-2 multispectral time series represented as tensors of shape $T \times C \times H \times W$, where $T$ denotes the number of temporal observations, $C$ the number of spectral bands, and $H \times W$ the spatial extent; in the single-pixel setting, 
($H = 1$, $W = 1$), 
while larger spatial dimensions enable the incorporation of additional spatial context. All Sentinel-2 bands are linearly rescaled to the range [0, 1] using per-band min–max normalization , where the minimum and maximum statistics are computed over the entire dataset. Learnable temporal positional embeddings of size 512 are added to each frame. The resulting sequence with CLS token is then fed into a 6-layer Transformer encoder with 8 attention heads, a hidden dimension of 256, and a latent size of 512. The Transformer employs GELU activations and LayerNorm throughout. The final class token output is passed through a lightweight two-layer MLP projection head to obtain the satellite embedding.

\paragraph{Contrastive Alignment.}
Both the CLIP-based ground embeddings and the satellite embeddings are projected into a shared 512-dimensional latent space prior to alignment. Both ground and satellite embeddings are L2-normalized and aligned via the contrastive objective described in Section~\ref{sec:method}. The momentum-based memory queue maintains $K=2048$ negative samples and is updated at each iteration through an enqueue–dequeue mechanism, where newly computed embeddings replace the oldest to keep the queue size constant.

\paragraph{Training Configuration.}
Models are trained using the AdamW optimizer with an initial learning rate of $10^{-4}$, weight decay of $1\times10^{-6}$, and $(\beta_1, \beta_2)=(0.9, 0.999)$. A cosine annealing schedule with 10 warm-up epochs is used over a total of 200 epochs, with a batch size of 1024. All experiments are conducted on a single NVIDIA TITAN X GPU. During training, positive pairs are formed between satellite and corresponding ground embeddings, while negatives are drawn from both the in-batch and the MoCo-style memory queue of size 2048.

\subsection{Evaluation Setup}
\label{sec:eval_setup}
The overall evaluation framework is illustrated in appendix Figure~\ref{fig:evaluation_setup}. During inference, only the trained satellite encoder is employed.  The evaluation encompasses three components that jointly assess the versatility and generalization of the proposed TimeSenCLIP framework across diverse remote sensing tasks:  
(1) \textbf{Zero-shot classification}, which measures the model’s ability to identify land cover, land use, crop type, and habitat classes purely from semantic text descriptions, without any task-specific fine-tuning;  
(2) \textbf{Cross-modal retrieval}, which evaluates the quality of representation alignment by computing similarity between embeddings from the satellite time-series and ground-level domains, enabling both Satellite-to-Ground (S2G) and Ground-to-Satellite (G2S) retrieval; and  
(3) \textbf{Scenicness prediction}, which leverages the learned multimodal embeddings to estimate perceptual “scenicness” scores, demonstrating the model’s capacity to generalize beyond categorical classification to subjective, aesthetic evaluation. More details on each evaluation tasks are in Appendix \ref{apx:eval_setup}.

In addition, we describe how temporal aggregations are applied to summarize multi-temporal inputs, outline the baseline models used for performance comparison, and specify the evaluation metrics adopted for each evaluation task.

\subsubsection{Temporal Aggregation} \label{sec:temp_agg}
In addition, during inference, we evaluate the model under three temporal aggregation strategies. The monthly setting uses all 12 individual time steps. The quarterly setting reduces the sequence to 4 time steps, each computed as the median of a 3-month window. The annual (Single) setting further compresses the sequence to a single representation by taking the median across all 12 months.

\subsubsection{Baselines.}
We compare TimeSenCLIP against a range of CLIP-based VLMs, including CLIP~\cite{radford2021learning}, GeoRSCLIP~\cite{zhang2023rs5m}, RemoteCLIP~\cite{liu2023remoteclip}, SkyCLIP~\cite{wang2024skyscript},  SenCLIP~\cite{jain2025senclip}, and Llama3-MS-CLIP~\cite{marimo2025beyond}. All baselines rely on spatial representations and operate on full $64\times64$ RGB image patches, with the exception of Llama3-MS-CLIP, which uses multispectral Sentinel-2 inputs that are resized to $224\times224$ to match the standard CLIP input resolution.

For a fair comparison, TimeSenCLIP is evaluated under multiple configurations:
(i) \textbf{TimeSenCLIP-P1-RGB}, which models single-pixel ($1\times1$) RGB time series and isolates the effect of temporal reasoning without spatial context;  
(ii) \textbf{TimeSenCLIP-P64-RGB}, which builds on SenCLIP-pretrained spatial embeddings extracted from $64\times64$ RGB patches and augments them with our temporal encoding module; and  
(iii) \textbf{TimeSenCLIP-P1-MS}, our main model, which leverages multispectral single-pixel ($1\times1$) time series to fully exploit both temporal and spectral information.

Each model is trained separately. This setup enables a direct comparison between spatially grounded VLMs and our temporally and spectrally enhanced representations, disentangling the contributions of pure temporal modeling (P1-RGB), temporally enriched spatial embeddings (P64-RGB), and temporally–spectrally informed multispectral inputs (P1-MS).

\subsubsection{Evaluation Metrics.}
To comprehensively assess the performance of our proposed framework, we employ task-specific evaluation metrics tailored to each downstream objective. The three evaluation settings are quantitatively evaluated as follows:

\begin{itemize}
\item Zero-shot classification:
We report the \textit{Top-1 accuracy}, which measures the proportion of test samples for which the predicted label (i.e., the class with the highest similarity score between the image and text embeddings) exactly matches the ground-truth class. This metric reflects the model’s direct classification capability without any fine-tuning or task-specific adaptation.
\item  Cross-Modal retrieval:  
Retrieval performance is quantified using \textit{Recall@1}, which indicates the percentage of queries where the top-ranked retrieved image corresponds to the same class as the query. This metric evaluates the discriminative quality of the learned embeddings and their alignment across modalities (e.g., Satellite-to-Ground and Ground-to-Satellite retrieval).

\item Scenicness regression: 
The predicted scenicness scores are evaluated using both the \textit{Pearson correlation coefficient ($R$)} and \textit{Kendall’s tau ($\tau$)}. Pearson’s $R$ measures the linear correlation between predicted and ground-truth scores, while Kendall’s $\tau$ assesses their rank correlation.
\end{itemize}

\section{Results and Discussion}

This section is organized into four parts. First, Section~\ref{sec:quantitative_eval} reports quantitative results for zero-shot classification, cross-modal retrieval, and scenicness regression. Second, Section~\ref{sec:ablation} analyses the effects of architectural design choices, spatial context, augmentation strategies, and computational efficiency through ablation studies. Third, Section~\ref{sec:qualitative} provides qualitative insights via per-pixel prediction maps and retrieval examples. Finally, Section~\ref{sec:discussion} contextualises these findings and discusses their broader technical and ecological implications.

\subsection{Quantitative Results}
\label{sec:quantitative_eval}
We evaluate TimeSenCLIP on three scenarios: zero-shot classification, cross-modal retrieval, and scenicness regression. These tasks collectively probe the model’s ability to generalize across domains, modalities, and temporal resolutions without task-specific fine-tuning. All experiments compare TimeSenCLIP against multiple CLIP-based baselines, enabling a consistent assessment of temporal, spectral, and spatial contributions.

\subsubsection{Zero-shot Classification Performance}
\label{sec:zs_clip}

\input{table_zeroshot_clipbased}

For zero-shot classification, we report Top-1 accuracy using three types of textual prompts: \emph{class-name}, \emph{generic}, and \emph{description-level}, as introduced in Section~\ref{subsec:prompts}. Evaluations are conducted across five geospatial tasks: Land Cover, Land Use, Habitat, Crops, and Bioregion, under varying temporal lengths ($T$), spatial input resolutions (\emph{Pix}), and spectral modalities (RGB and multispectral). 
Table~\ref{tab:zeroshot_clip} summarizes the zero-shot classification performance across all baseline models and TimeSenCLIP variants across these settings. Each experiment was repeated with five random seeds; column values show mean performance, while the final column reports the average standard deviation.

Standard CLIP models show limited transfer to Sentinel-2 imagery due to the lack of temporal cues and the domain gap between natural and multispectral satellite images. GeoRSCLIP, and SenCLIP, improve upon standard CLIP but still struggle to capture intra-annual variations, leading to inconsistent performance across tasks and prompt types.  Llama3-MS-CLIP further improves over RGB-based baselines for Land Cover and Land Use, suggesting that multispectral inputs help capture broad material and surface properties. However, its performance degrades on more complex tasks such as Crops, Habitat, and Bioregion.

TimeSenCLIP consistently outperforms all prior CLIP-based approaches across tasks, temporal settings, and prompt formulations.
Notably, \textbf{TimeSenCLIP-P1-RGB}, trained solely on single-pixel RGB time series, surpasses all existing CLIP baselines, demonstrating that temporal modeling alone can capture discriminative phenological patterns even without spatial context.
This effect is especially pronounced for temporally driven tasks such as Crops and Bioregion.

Spatial context remains beneficial when temporal coverage is limited.
\textbf{TimeSenCLIP-P64-RGB}, which combines SenCLIP-pretrained spatial embeddings with a temporal transformer, achieves strong performance on spatially structured tasks such as Land Use and Land Cover.
However, as temporal depth increases, single-pixel models often match or exceed their larger-patch counterparts, indicating that temporal evolution can outweigh spatial detail for many geospatial categories.

Incorporating multispectral inputs further enhances performance.
\textbf{TimeSenCLIP-P1-MS} consistently improves over RGB variants, particularly for Land Cover and Crops, highlighting the complementary role of spectral information when combined with temporal modeling.
Overall, these results confirm that temporal modeling is the dominant factor for zero-shot generalization in remote sensing, with spatial and spectral cues providing task-dependent gains.

Noticeably, Crops and Bioregion benefit most from temporal information due to strong seasonal signatures.
Land Cover and Land Use gain moderate improvements from combining spatial and temporal cues.
In contrast, Habitat classification shows limited sensitivity to increased temporal depth, suggesting that habitat-level semantics are driven primarily by spectral composition and spatial structure rather than seasonal evolution.

\subsubsection{Cross-Modal Retrieval Performance}

\input{table_ItoI_clipbased}
For cross-modal retrieval, we evaluate cross modality matching performance using Recall@1 in both Ground-to-Satellite (G2S) and Satellite-to-Ground (S2G) directions. Experiments are conducted across six semantic tasks: Land Cover, Land Use, Habitat, Crops, Bioregion, and Country, while varying temporal length ($T$), spatial input resolution (\emph{Pix}), and spectral modality (RGB versus multispectral). The results compare all CLIP-based baselines and TimeSenCLIP variants under consistent retrieval protocols.

The cross-modal retrieval results in Table~\ref{tab:retrieval_results_clip} reveal the same fundamental trends observed for the zero-shot classification scenario. Standard CLIP and SkyCLIP achieve only moderate performance and exhibit a strong degradation in Satellite-to-Ground (S2G) retrieval, reflecting limited robustness to viewpoint, scale, and acquisition-time discrepancies. GeoRSCLIP and SenCLIP improve retrieval for structurally stable categories such as Land Cover and Land Use, yet remain sensitive to seasonal variability, leading to weaker performance on temporally dynamic tasks such as Crops. Multispectral adaptation in Llama3-MS-CLIP yields only marginal retrieval gains.

In contrast, TimeSenCLIP delivers the strongest and most stable cross-modal alignment across tasks and retrieval directions, with performance improving consistently with temporal depth and peaking on phenology-driven categories, highlighting temporal dynamics as a view-invariant signal for matching ground and satellite observations. Although larger spatial patches remain beneficial for coarse or regionally defined tasks, temporally rich single-pixel multispectral sequences often achieve comparable or superior performance, indicating that temporal–spectral information can compensate for reduced spatial context. Moreover, TimeSenCLIP maintains balanced accuracy between Ground-to-Satellite and Satellite-to-Ground retrieval, unlike conventional CLIP-based models that exhibit pronounced asymmetry, demonstrating that temporal encoding produces stable, view-agnostic representations that support consistent semantic alignment across perspectives.

\subsubsection{Scenicness Regression with Prompt Ensembling}
\label{subsec:scenicness}

\input{table_scenicness}

We evaluate scenicness prediction using prompt-ensembling strategies and report performance with Pearson’s $R$ and Kendall’s $\tau$. Table~\ref{tab:scenicness_main} compares early and late prompt ensembling approaches for scenicness prediction following~\cite{levering2024prompt}, reporting performance on Sentinel-2 imagery using both correlation metrics. More details on setup are in Appendix \ref{apx:eval_setup}

Table~\ref{tab:scenicness_main} shows that with single pixel temporal depth is critical for perceptual landscape estimation.  With a single timestamp ($T{=}1$), performance is limited ($R{=}0.318$, $\tau{=}0.205$ under late ensembling), indicating that spectral information alone is insufficient. However, increasing temporal coverage yields consistent improvements, with quarterly aggregation enhancing performance and monthly temporal input ($T{=}12$) achieving the best results under single pixel ($R{=}0.527$, $\tau{=}0.363$).  

CLIP continues to serve as a competitive non-temporal baseline, but its static representations ($R$ = 0.517, $\tau$ = 0.362) fall short of TimeSenCLIP-P64-RGB, demonstrating that temporal cues are valuable even for tasks commonly considered visually “static”. Under single-temporal input, TimeSenCLIP-P64-RGB achieves the strongest performance ($R$ = 0.660, $\tau$ = 0.450), and also shows competitive performance with CLIP on ground-level imagery ($R$ = 0.660, $\tau$ = 0.465). These results highlight two key aspects of our design:
(1) temporal encoding substantially enhances the SenCLIP spatial encoder, equipping it to capture seasonal cycles and long-term landscape appearance; and
(2) large spatial context (P64) remains important for modeling holistic perceptual attributes such as scenicness, where multi-scale texture, landform geometry, and spatial composition strongly influence aesthetic ratings.

Additional detailed results are provided in Appendix~\ref{apx:scennicness_full}.

\subsection{Ablation Study}
\label{sec:ablation}
\input{table_zeroshot_baseline}

\subsubsection{Conventional Temporal Baseline Comparison}

For a fair comparison with TimeSenCLIP's transformer-based encoder, we train classic temporal models CNN1D \cite{kiranyaz2015real}, MLP \cite{rumelhart1986learning}, ConvTran, and TempCNN, following the same training protocol as TimeSenCLIP and using identical multispectral single-pixel time series inputs. 

These architectures are largely adopted for the classification of satellite image time series data and provide complementary inductive biases: MLPs capture per-timestep spectral relationships, CNN1D models learn local temporal patterns, TempCNN extends this with deeper temporal convolutions, and ConvTran introduces hybrid convolution–transformer reasoning. 
By restricting all baselines to purely temporal–spectral signals and removing spatial context, we ensure that performance differences reflect the models' temporal modeling capabilities rather than advantages from spatial structure.

\label{sec:zs_classical}

Table~\ref{tab:zeroshot} reports the zero-shot evaluation results, with each experiment repeated over five random seeds. Reported values correspond to the mean performance for each task, while the final column summarizes the average standard deviation across runs. The results indicate that classic time series architectures achieve competitive zero-shot performance, but their effectiveness varies notably across model families and task types. MLP and TempCNN generally provide stronger results than CNN1D and ConvTran, particularly on tasks where spectral or phenological cues play a larger role. TempCNN achieves the highest scores on couple of benchmarks, especially for Land Cover and Habitat, reflecting the strength of convolutional temporal filters for modeling smooth spectral-temporal trends. MLP remains consistently stable across tasks, offering strong performance despite its lightweight structure. CNN1D and ConvTran deliver more modest accuracy overall, showing limitations in capturing the complexity of multispectral sequences in a zero-shot setting.

TimeSenCLIP’s transformer variant performs competitively across all tasks and often best performing models, particularly for Crops and Bioregions, where its learned representations capture class distinctions that benefit from subtle spectral-temporal differences. Although differences between architectures are sometimes limited, the transformer-based approach demonstrates balanced behavior across all prompt types Class, Generic, and Description, resulting in the highest mean Overall accuracy under all types of prompts. This suggests that the combination of our training strategy and transformer design produces embeddings that generalize well across diverse ecosystems and LULC categories without reliance on spatial context or tailored text supervision.

The supervised upper bound provides the highest overall accuracy, but the performance gap between supervised and zero-shot models varies substantially across tasks. Smaller gaps for Land Cover and Land Use suggest that these broad categories are reasonably well represented in the CLIP text encoder, enabling effective semantic alignment even without task-specific training. In contrast, larger discrepancies appear for Habitat Crops, and Bioregions. This can be attributed not only to limitations in the pretrained text encoder, which offers less coverage of specialized ecological terminology, but also to the intrinsic difficulty of these tasks: they involve a much larger number of classes (e.g., 44 habitat types and 39 crop types) in contrast to land cover and land use (e.g., 8 land cover types and 16 land use types), making semantic separation in the embedding space more challenging. Additionally, many classes in these domains exhibit weak or overlapping phenological and ecological signatures (e.g., similar crop growth cycles, or habitats without distinct seasonal patterns), reducing the discriminative cues available for zero-shot inference.

\subsubsection{Impact of Spatial Context}
\begin{figure*}
\centering
\includegraphics[width=\textwidth]{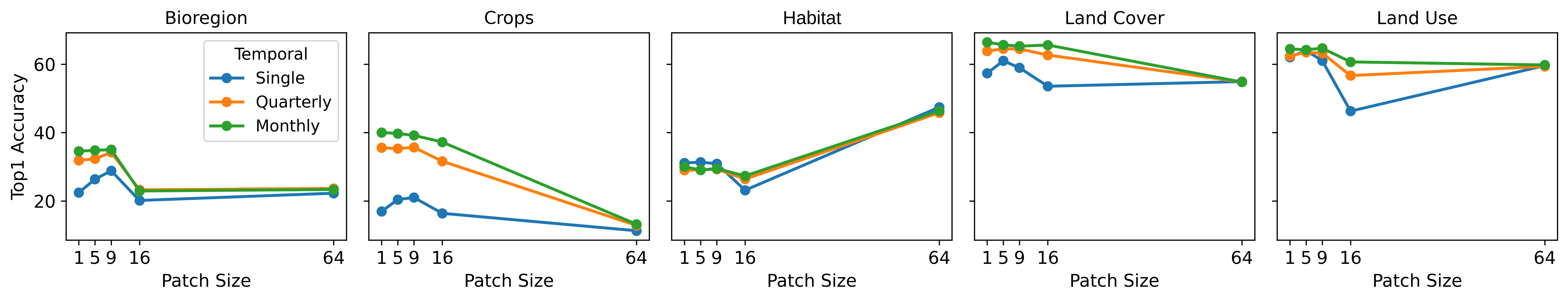}
\caption{Impact of spatial information on zero-shot classification Top-1 accuracy (\%) across different mapping tasks. Single-pixel (P1) time series are often competitive with larger patches.}
\label{fig:spatial_zeroshot}
\end{figure*}

We investigate how spatial context, controlled via patch size, affects zero-shot classification performance. Specifically, we compare single-pixel (P1), $5 \times 5$ (P5), $9 \times 9$ (P9), $16 \times 16$ (P16), and $64 \times 64$ (P64) patches, where larger patches provide broader spatial context to the temporal encoder. A separate model is trained for each patch size and then evaluated on corresponding size and with different temporal settings. Figure~\ref{fig:spatial_zeroshot} illustrates the impact of patch size on Top-1 accuracy across several geospatial classification tasks.

The results reveal that single-pixel (P1) time series are highly competitive, often matching or exceeding larger patches. This demonstrates that temporal and spectral dynamics alone capture much of the discriminative information required for zero-shot classification, with minimal spatial context being sufficient for robust geospatial representations. Small spatial neighborhoods (P5) provide modest gains for some tasks; for example, Land Cover zero-shot accuracy increases from 56.08\% (P1) to 58.14\%, indicating that limited spatial information can complement temporal and spectral cues. Larger patches (P16–P64) mainly benefit coarse or spatially homogeneous tasks such as Habitat or Land Cover/Use, whereas fine-grained categories like Crops or Bioregion can degrade due to mixed-signal pixels.

The temporal depth consistently dominates over spatial size. Increasing the number of timestamps improves zero-shot accuracy across all patch sizes, and long P1 time series can match or surpass larger patch variants, especially for phenology-driven tasks. These results highlight that minimal spatial input combined with sufficient temporal coverage provides a strong, generalizable geospatial representation while keeping computational costs low.

\subsubsection{Impact of Dropout Strategies}
To assess how TimeSenCLIP handles incomplete or noisy observations, common in real-world remote sensing, we study two dropout-based regularization strategies on $T=12$ temporal setting, applied only during training:
\begin{itemize}
    \item Temporal Drop (TS Drop): for each batch, with a probability of 50\%, all timesteps of the time series are either randomly masked, quarterly masked or collapsed into a single median value as discussed in Section~\ref{sec:temp_aug}. This simulates missing temporal observations or coarsely aggregated products.
    \item Multispectral Dropout (MS Drop): for each batch, non-RGB spectral bands are randomly masked with a probability of 50\%, encouraging the model to rely on robust spectral cues.
\end{itemize}
During evaluation, we assess the robustness of models trained with dropout under different temporal aggregations: Single and Monthly, as described in Section~\ref{sec:temp_agg}.  

\input{table_dropout_zeroshot}

Zero-shot classification results in Table~\ref{tab:zeroshot_drops} indicate that temporal dropout is highly effective when temporal coverage is limited. In the single-temporal setting, TS Drop substantially improves performance, increasing the average Top-1 accuracy from 17.2\% to 36.7\%. Without temporal augmentation, the model tends to overfit to the fixed temporal signature of a single acquisition. TS Drop forces the model to learn more stable, temporally robust cues, demonstrating that temporal regularization is particularly valuable for sparse time series.

In contrast, MS Drop provides minimal gains in some settings but often reduces performance. Removing non-RGB bands discards spectral features critical for distinguishing vegetation and ecosystem types, limiting its usefulness for zero-shot classification. Combining TS Drop with MS Drop does not produce additional improvements, suggesting that the primary source of robustness comes from temporal augmentation rather than spectral masking.

These findings highlight temporal dropout as a simple yet powerful strategy for improving TimeSenCLIP’s generalization under limited temporal observations. As temporal depth increases, the model naturally develops temporal robustness, and the relative benefit of TS Drop diminishes.

\subsubsection{Computational Efficiency and Inference Performance}
\label{sec:performance_analysis}
Table \ref{tab:model_perf} summarizes the computational efficiency of the proposed TimeSenCLIP model in comparison to CLIP-based baselines. All CLIP-based models behave identically in terms of FLOPs, parameters, and throughput, as they share the same image encoder. TimeSenCLIP, which processes 12 temporal frames with 10 spectral bands, achieves a ~97–98\% reduction in FLOPs and ~94\% fewer parameters compared to CLIP when operating on a single pixel (1×1), while delivering substantially higher throughput (over 33× faster) and requiring significantly less GPU memory. Even when processing larger spatial patches such as 5×5, 9×9, or 16×16, TimeSenCLIP maintains low computational cost, with only modest increases in FLOPs, memory, and throughput. These results demonstrate that TimeSenCLIP is highly efficient and well-suited for scalable remote sensing tasks, particularly in resource-constrained or real-time settings. As TimeSenCLIP-P64 uses SenCLIP for embeddings, so it was not included in this comparison.

\begin{table}[ht]
\centering

\resizebox{\columnwidth}{!}{%
\begin{tabular}{lcccc}

\toprule
\textbf{Model}  & \textbf{\begin{tabular}[c]{@{}c@{}}FLOPs\\ (GMac)\end{tabular}} & \textbf{\begin{tabular}[c]{@{}c@{}}Parameters\\ (M)\end{tabular}} & \textbf{\begin{tabular}[c]{@{}c@{}}Throughput\\ (samples/s)\end{tabular}} & \textbf{\begin{tabular}[c]{@{}c@{}}Peak Memory \\ (MB)\end{tabular}} \\ 
\midrule
CLIP     & 4.46   & 151.28  & 565.33   & 3045.13 \\
TimeSenCLIP 1x1    & 0.105  & 8.17   & 19530.50 & 584.60  \\
TimeSenCLIP 5x5    & 0.106 & 8.29    & 19424.24 & 596.32 \\
TimeSenCLIP 9x9    & 0.110  & 8.58    & 19428.41 & 624.08 \\
TimeSenCLIP 16x16  & 0.121  & 9.48    & 17798.35 & 709.12 \\
\bottomrule
\end{tabular}}
\caption{Computational efficiency of CLIP-Based Model and TimeSenCLIP models on Batch Size 1024.}
\label{tab:model_perf}
\end{table}

\subsection{Qualitative Analysis}
\label{sec:qualitative}
\subsubsection{Per-Pixel Prediction Maps}

\begin{figure*}
    \centering
    \caption{Per-pixel land-cover prediction map generated from text-prompt-based LULC class descriptions using monthly Sentinel-2 time series data (64×64-pixel tile). The accompanying Bing aerial image provides a high-resolution visualization of the same geographic area for reference only and is centered on the corresponding Sentinel-2 tile. As the two datasets are acquired at different times and resolutions, discrepancies in appearance and feature alignment are expected. The color bar indicates the set of LULC classes predicted within the tile. The actual class for this geo-tagged location corresponds to the habitat ground-truth label as they provide broader class context for larger spatial than LUCAS LULC labels.}
    \includegraphics[width=\linewidth]{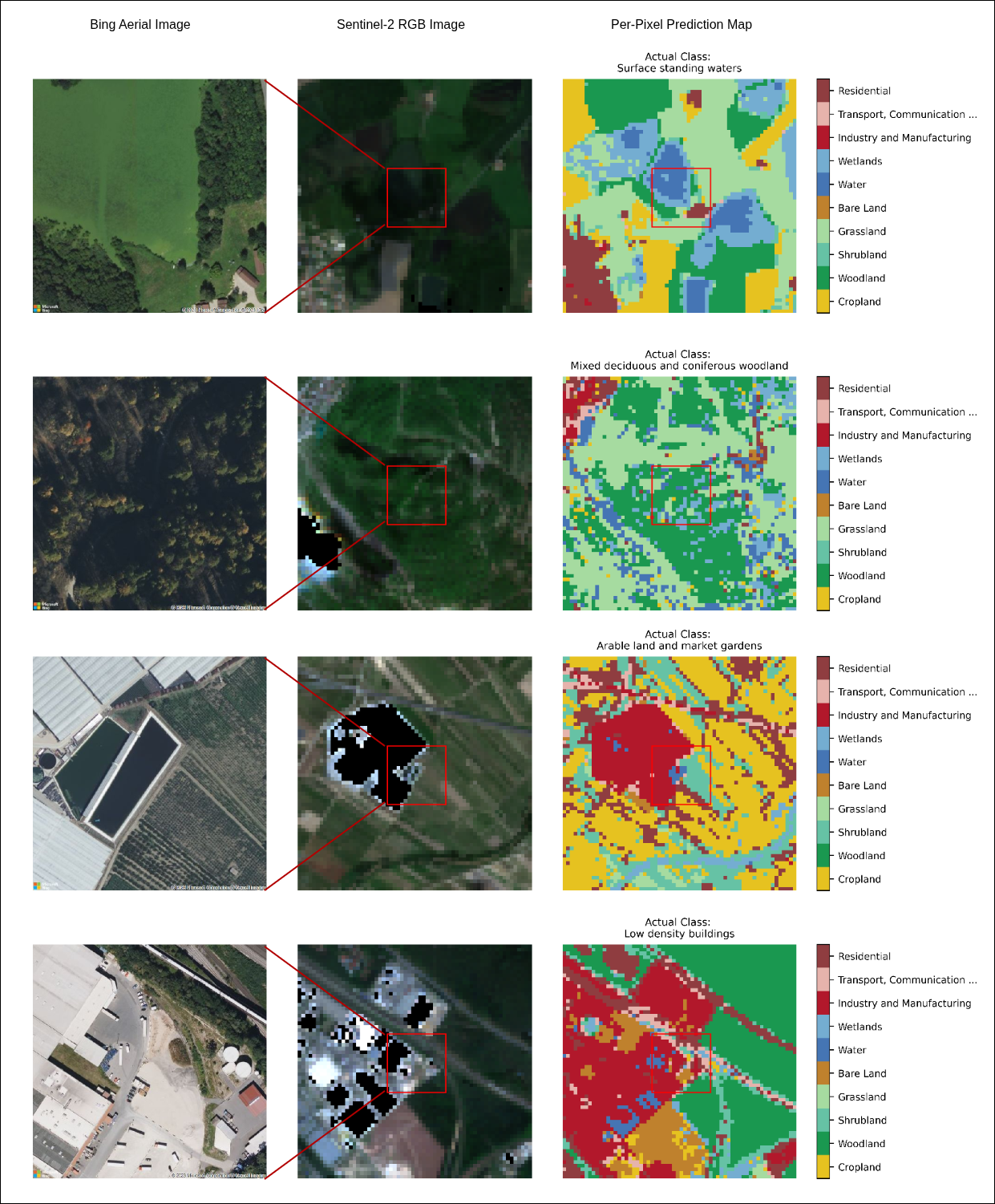}
    
    \label{fig:pred_maps_lulc}
\end{figure*}

\begin{figure*}
    \centering
     \caption{Per-pixel crop type prediction map generated using text-prompt-based crop class descriptions on monthly Sentinel-2 time series (64×64-pixel tile). The Bing aerial image provides a high-resolution visual reference of the same region, centered on the Sentinel-2 tile. Because the Bing and Sentinel-2 images were acquired at different times, visual appearance may differ; the Bing image is included solely as contextual reference. The color bar indicates the set of crop classes predicted within this tile. The actual class for this geo-tagged location corresponds to the LUCAS ground-truth label. To improve interpretability, fine-grained crop classes were color-coded according to their higher-level agronomic groups.}
    \includegraphics[width=\linewidth, scale=0.2]{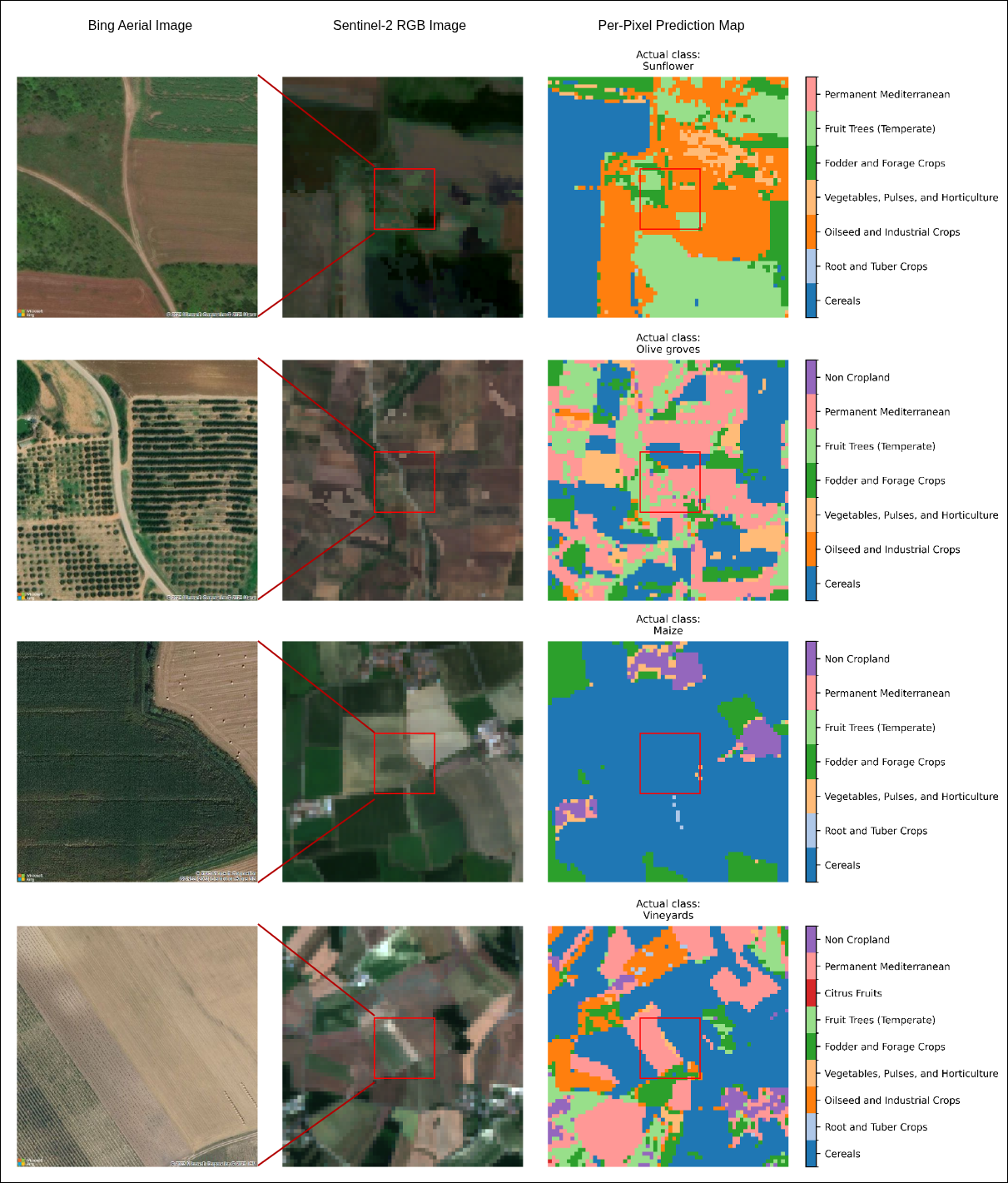}
   
    \label{fig:pred_maps_crops}
\end{figure*}

To evaluate the capability of TimeSenCLIP to perform per-pixel zero-shot classification using natural-language class descriptions, we derive prediction maps for LULC and crop type categories. We report such maps in Figure \ref{fig:pred_maps_lulc} and Figure \ref{fig:pred_maps_crops}, together with reference ground-truth information to facilitate qualitative comparison. Because dense semantic segmentation annotations are not available in LUCAS, only the class label at the center pixel is provided; this central label is therefore shown as the ground-truth reference rather than a full pixel-wise segmentation map. Each extract spans $640m \times 640m$ (corresponding to a $64 \times 64$ Sentinel-2 pixel tile) and represents a monthly multispectral composite. Classifications are produced at the pixel level using the single-pixel TimeSenCLIP model, which analyzes each pixel time series independently and performs zero-shot classification using descriptive text prompts (Section~\ref{subsec:prompts}). Each pixel is assigned to the class whose textual description best matches its spectral–temporal signature, following the same zero-shot procedure described in Sections~\ref{sec:zs_clip} and~\ref{sec:zs_classical}, but applied densely across the $64 \times 64$ tile.

The maps exhibit spatially coherent and semantically meaningful patterns, illustrating the model’s ability to associate high-level language-based class concepts with pixel-scale temporal behavior. At the same time, localized salt-and-pepper noise appears as isolated pixel-level fluctuations between neighboring classes. This effect is expected because inference is performed pixel-wise and no spatial smoothing post-processing is used. Nevertheless, broader spatial structures remain consistent, indicating that spectral–temporal representations capture meaningful land-cover organization despite the absence of explicit spatial modeling.

To facilitate visual interpretation, we include a high-resolution Bing aerial image for each location as a reference. Although the Bing imagery is not perfectly aligned with Sentinel-2 due to differences in acquisition date, spatial resolution, and viewing geometry, it still provides useful contextual cues regarding settlement structure, vegetation distribution, and general land-use patterns. The LUCAS point label associated with each location is also shown to indicate the ground-truth class at the sample center, helping assess the semantic correctness of the predicted patterns.

For LULC zero-shot pixel classification, we merged land-use and land-cover taxonomies and resolved overlapping categories by unifying semantically equivalent classes. For example, the land-cover class Cropland and the land-use class Agriculture were consolidated into a single category, Cropland. As shown in Figure~\ref{fig:pred_maps_lulc}, the model correctly identifies a wide range of LULC categories and, in ambiguous cases, assigns semantically consistent or closely related classes, including built-up classes such as residential and industrial areas, as well as natural categories such as grassland and shrubland.

For zero-shot crop pixel classification, we used the full set of fine-grained crop class descriptions. After generating prediction maps at the fine-grained level, classes were aggregated into higher-level agronomic groups for visualization. For example, Common wheat, Maize, and Barley were grouped under Cereals, while Olive groves and Vineyards were grouped as Permanent Mediterranean. As shown in Figure~\ref{fig:pred_maps_crops}, although certain crop categories are over-predicted, likely reflecting the difficulty of zero-shot discrimination between spectrally similar crops; the model consistently identifies the correct crop class at the central LUCAS location and effectively distinguishes cropland from artificial surfaces or non-cropland areas.

\subsubsection{Text-To-Image Retrieval}
\begin{figure*}
    \centering
     \caption{Text-to-image retrieval performance using TimeSenCLIP with descriptive prompts. While the model processes single pixel monthly temporal composites during operation, these visualizations show representative RGB 64×64 patches for interpretability. Results demonstrate the model's ability to match textual descriptions with relevant satellite imagery across diverse land cover, land use, bioregions, habitat and crops classes. Each row presents the Top-5 retrieved time-series with their corresponding class labels and bioregion. Class labels are drawn from LULC, Habitat and Crops classes, when a crop label is unavailable, it is denoted with ``-" . }
    \includegraphics[scale=0.5]{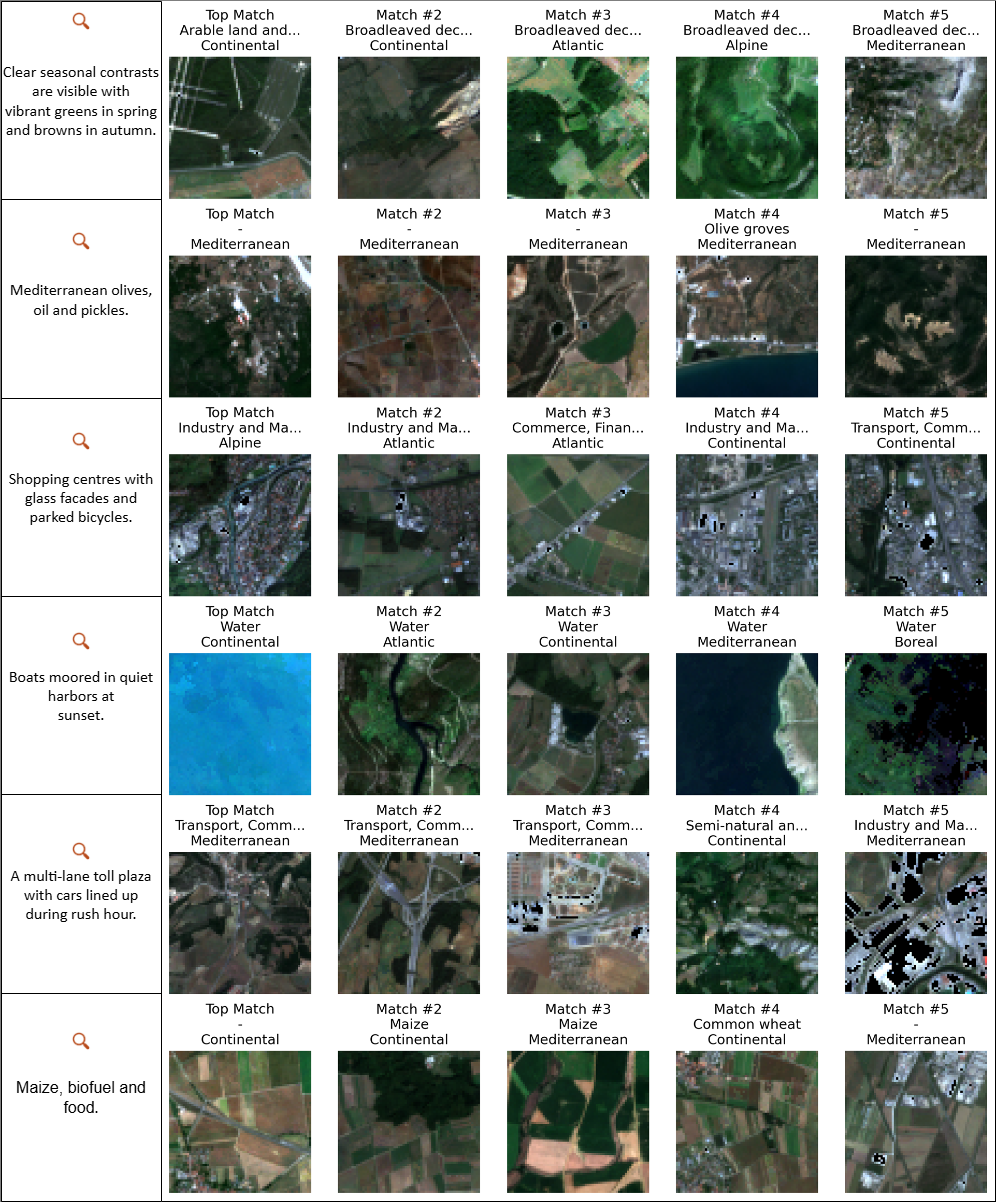}
    
    \label{fig:texttoimage}
\end{figure*}

Figure~\ref{fig:texttoimage} presents the quantitative results of text-to-image retrieval, where free-form textual descriptions are used to search for the most semantically relevant satellite image patches. Each query corresponds to a specific land cover, land use, habitat class, biogeographic region, or crop type.

Our model, TimeSenCLIP, demonstrates strong performance in matching descriptive text prompts to appropriate satellite imagery, despite the fact that these descriptions are often abstract, high-level, or grounded in human perception (e.g.,`\textit{`traditional olive landscapes in Greece, Italy, Spain."}).

For each text query, we display the Top-5 retrieved satellite patches. These examples showcase the model's ability to generalize across tasks and semantic levels, from broad biogeographic zones to fine-grained crop types. Retrieved visually and semantically consistent matches, indicating that the learned representation aligns textual semantics with satellite spectral-temporal features effectively.

This experiment demonstrates the practical utility of TimeSenCLIP in real-world scenarios, where users can retrieve relevant satellite imagery using natural language descriptions without relying on predefined class labels. It also underscores the model's strong zero-shot generalization capabilities, as the textual prompts and categories used during retrieval were not part of the training process. Instead, the model learns to associate semantic meaning through supervision from ground-level imagery, enabling it to match unseen descriptions to appropriate satellite scenes.

\subsection{Discussion}
\label{sec:discussion}
Temporal dynamics play an important role in learned feature representations from remote sensing data, providing the strongest discriminative signal across zero-shot classification and cross-view retrieval, particularly for phenology-driven categories such as Crops and Bioregions where static spatial appearance is not informative enough. When temporal frequency is dense, single-pixel time series often match or exceed larger spatial patches, indicating that temporal information can partially compensate for limited spatial context and challenging the assumption that broader spatial extent is always required for geospatial understanding. Nevertheless, the relative contribution of temporal and spatial cues is task-dependent: Crops and Bioregions are primarily influenced by seasonal dynamics, Land Cover and Land Use benefit from spatial clues, and Habitat shows limited sensitivity to temporal depth, reflecting its reliance on fine spatial structure or additional semantic priors.

Ablation studies provide ecological insight, showing that time series dynamics alone capture phenologically distinct ecosystem signatures, with ecosystem identity expressed through temporal–spectral patterns rather than spatial structure. Higher accuracy for bioregions than finer crop or habitat classes reflects the greater temporal stability of large-scale ecological systems, while robustness to temporal dropout suggests ecosystems differ in resilience to seasonal gaps. Phenology thus serves as a proxy for ecosystem function, enabling ecological mapping with limited spatial context~\cite{lausch2016linking}.

Multispectral inputs enhance these temporal representations, improving the separability of visually similar classes with distinct spectral–temporal dynamics, especially for Land Cover and Crops. Textual discriptive prompts also provide complementary guidance, injecting domain knowledge not readily observable from imagery alone and reinforcing the importance of prompt design in zero-shot remote sensing models.

Finally, these performance gains are achieved with substantial computational efficiency. TimeSenCLIP’s temporal transformer processes multi-spectral time series for single pixels or small patches with dramatically lower FLOPs, memory, and inference time compared to standard CLIP-based architectures, while maintaining or exceeding accuracy. This efficiency enables scalable, high-throughput analysis of large satellite datasets, highlighting a favorable trade-off between representational richness and computational cost.
\section{Conclusion}
This work presents TimeSenCLIP, a temporally and spectrally-aware multimodal framework for zero-shot ecological classification and retrieval. By leveraging fine-grained spectral-temporal inputs and regularization strategies, TimeSenCLIP demonstrates consistent and substantial improvements over existing baselines across diverse remote sensing tasks, including Land Cover, Land Use, habitat mapping, Bioregions, Crop types, and Scenicness prediction. Our findings highlight the importance of taking into account temporal dynamics particularly at a monthly resolution to capture seasonal and phenological patterns that are critical for fine-grained geospatial understanding.

Ablation studies confirm that temporal dropout enhances robustness to missing or inconsistent observations, while spatial context offers task-specific benefits. Notably, TimeSenCLIP maintains strong performance even when constrained to single pixel inputs, underscoring the power of spectral-temporal learning via contrastive learning. For tasks such as Crop type classification and Scenicness prediction, temporal ensembling and prompt diversification further improve performance, illustrating the value of aligning learned representations with temporal variations. 

\paragraph{Limitations and Future Work}

Although our model shows strong performance on ecological tasks in Europe, it has only been evaluated using Sentinel-2 and LUCAS data from the EU. This limits the understanding of its generalizability to other sensors and regions with different climates and phenologies. Extending the framework to manage multi-sensor information (e.g. Sentinel-1 and Sentinel-2) and scale up globally by combining multiple sources of ground-level photographs are key directions to improve our approach.
Another inherent limitation resides in the CLIP representation itself. Being based on the alignment between internet photographs and their captions, many characteristics that are only observable across time may not be captured by the CLIP image embeddings, and cannot, thus, be transferred to TimeSenCLIP.
This could be achieved by complementing the ground-level image representations with textual geocoded descriptions designed to fill these gaps.

\noindent \textbf{Acknowledgements}\\
This research was supported by the ‘Giving Rural Actors Novel Data and Re-Usable Tools to Lead Public Action in Rural Areas’ (GRANULAR)
project, which has received funding from the European Union’s Horizon Europe Research and Innovation Programme under Grant Agreement No.
101061068.
This work was also supported in part by the ANR project OBTEA (ANR-22-CPJ1-0054-01) and I-SITE Excellence Program of the University of Montpellier, under the Investissements France 2030.

{
	\begin{spacing}{1.17}
		\normalsize
		\bibliography{ISPRSguidelines_authors} 
	\end{spacing}
}

\input{appendix}

\end{document}

%% file: table_prompt_example.tex
\begin{table*}[ht]
    \centering
    \footnotesize
    \setlength{\tabcolsep}{4pt}
    
    \resizebox{\textwidth}{!}{%
    \begin{tabular}{
        >{\centering\arraybackslash}m{0.25\textwidth} 
        m{0.75\textwidth}
    }  
    \hline
    \textbf{Category} & \textbf{Descriptive Prompts} \\ 
    \hline

    Sparsely wooded grasslands (EUNIS) &  
    Sparse canopy with open fields; scattered trees on grass; visible bare ground between patches. \\
    \hline

    Seasonally wet and wet grasslands (EUNIS) &  
    Wet meadows with reflective surfaces; dark wet zones amid dry land; green patches with water. \\
    \hline

    Boreal (Bio-Region) &  
    Dense pine and spruce forests; mossy forest floors in coniferous areas. \\
    \hline

    Mediterranean (Bio-Region) &  
    Olive/vineyard rows on slopes; dry, dusty terrain with sparse summer vegetation. \\
    \hline

    Common Wheat (Crops) &  
    Wheat fields near harvest; thriving wheat across Europe. \\
    \hline

    Olive Groves (Crops) &  
    Mediterranean olives, oil and pickles.; traditional olive landscapes in Greece, Italy, Spain. \\
    \hline

    Scenicness &  
    Wide motorway; busy highway; misty mountain lake; grassy bog surrounded by hills. \\
    \bottomrule

    \end{tabular}%
    }
    \caption{Examples of descriptive prompts, together with their associated category.}
    \label{tab:prompt_example}
\end{table*}

%% file: table_data_stats.tex
\begin{table*}[htbp]

\centering

\resizebox{\linewidth}{!}{%
\begin{tabular}{l c p{4cm} r r || l c p{4cm} r r}
\hline
\textbf{Label Type} & \# Classes & \textbf{Top 8 Classes (by proportion)} & \# Samples & Proportion (\%) & \textbf{Label Type} & \# Classes & \textbf{Top 8 Classes (by proportion)} & \# Samples & Proportion (\%) \\ \hline
Land Cover & 8 & Woodland & 17,813 & 35.48 & EUNIS Ecosystem & 44 & Arable land and market gardens & 15,732 & 31.34 \\
           &   & Cropland & 13,007 & 25.91 &                 &    & Broadleaved deciduous woodland & 8,228 & 16.39 \\
           &   & Grassland & 10,954 & 21.82 &                 &    & Coniferous woodland & 6,496 & 12.94 \\
           &   & Artificial Land & 3,173 & 6.32 &                 &    & Mesic grasslands & 6,197 & 12.34 \\
           &   & Shrubland & 2,715 & 5.41 &                 &    & Mixed deciduous and coniferous woodland & 2,549 & 5.08 \\
           &   & Bare Land & 1,139 & 2.27 &                 &    & Buildings of cities, towns and villages & 1,778 & 3.54 \\
           &   & Wetlands & 1,009 & 2.01 &                 &    & Low density buildings & 1,533 & 3.05 \\
           &   & Water & 395 & 0.79 &                 &    & Dry grasslands & 1,091 & 2.17 \\ \hline
Land Use   & 16 & Agriculture & 22,309 & 44.44 & Bio Region & 8 & Continental & 15,642 & 31.16 \\
           &    & Forestry & 15,081 & 30.04 &             &   & Mediterranean & 11,609 & 23.12 \\
           &    & Semi-natural and natural areas not in use & 6,806 & 13.56 & &   & Atlantic & 10,287 & 20.49 \\
           &    & Residential & 2,065 & 4.11 &             &   & Boreal & 7,320 & 14.58 \\
           &    & Transport, Communication Networks, Storage, Protection Works & 1,860 & 3.70 & &   & Alpine & 3,714 & 7.40 \\
           &    & Recreation, Leisure, Sport & 596 & 1.19 & & & Pannonian & 1,195 & 2.38 \\
           &    & Abandoned areas & 403 & 0.80 & & & Steppic & 331 & 0.66 \\
           &    & Community services & 346 & 0.69 & & & Black Sea & 107 & 0.21 \\ \hline
Crops      & 40 & Common wheat & 2,422 & 4.82 & Countries & 28 & France & 7,219 & 14.38 \\
           &    & Maize & 1,716 & 3.42 &           &    & Spain & 6,547 & 13.04 \\
           &    & Barley & 1,277 & 2.54 &           &    & Italy & 4,195 & 8.36 \\
           &    & Rape and turnip rape & 780 & 1.55 &   &    & Germany & 4,016 & 8.00 \\
           &    & Sunflower & 402 & 0.80 &           &    & Sweden & 4,000 & 7.97 \\
           &    & Oats & 364 & 0.73 &               &    & Poland & 3,463 & 6.90 \\
           &    & Durum wheat & 361 & 0.72 &          &    & United Kingdom & 2,582 & 5.14 \\
           &    & Rye & 338 & 0.67 &                 &    & Romania & 2,507 & 4.99 \\ \hline
\end{tabular} }

\caption{Class distribution statistics for the evaluation datasets across six label types. The table lists the total number of classes and the eight most frequent classes by proportion.}
\label{tab:data_stats}
\end{table*}

%% file: table_zeroshot_clipbased.tex
\begin{table*}[t]
\centering
\small
\setlength{\tabcolsep}{3.5pt}
\renewcommand{\arraystretch}{1.15}

\resizebox{\textwidth}{!}{

\begin{tabular}{lllcccccccccccccccccc}
\hline
\textbf{T} &
  \textbf{Bands} &
  \textbf{Model} &
  \multicolumn{3}{c|}{\textbf{Land Cover}} &
  \multicolumn{3}{c|}{\textbf{Land Use}} &
  \multicolumn{3}{c|}{\textbf{Habitat}} &
  \multicolumn{3}{c|}{\textbf{Crops}} &
  \multicolumn{3}{c|}{\textbf{Bioregion}} &
  \multicolumn{3}{c}{\textbf{Overall}} \\ \cline{4-21} 
\multicolumn{1}{c}{} &
  \multicolumn{1}{c}{} &
  \multicolumn{1}{c}{} &
  C &
  G &
  \multicolumn{1}{c|}{D} &
  C &
  G &
  \multicolumn{1}{c|}{D} &
  C &
  G &
  \multicolumn{1}{c|}{D} &
  C &
  G &
  \multicolumn{1}{c|}{D} &
  C &
  G &
  \multicolumn{1}{c|}{D} &
  C &
  G &
  D \\ \hline
\multicolumn{21}{l}{\textbf{64x64 Pix}} \\ 
\hline
\multirow{7}{*}{1} &
  RGB &
  CLIP &
  30.26 &
  29.40 &
  \multicolumn{1}{c|}{33.82} &
  40.20 &
  46.35 &
  \multicolumn{1}{c|}{38.38} &
  11.71 &
  14.89 &
  \multicolumn{1}{c|}{20.61} &
  3.03 &
  2.52 &
  \multicolumn{1}{c|}{4.10} &
  16.64 &
  17.55 &
  \multicolumn{1}{c|}{16.38} &
  20.37$\pm$.22 &
  22.14$\pm$.21 &
  22.66$\pm$.20 \\
 &
  RGB &
  GeoRSCLIP  &
  35.18 &
  34.66 &
  \multicolumn{1}{c|}{42.79} &
  35.58 &
  47.16 &
  \multicolumn{1}{c|}{46.87} &
  13.98 &
  18.15 &
  \multicolumn{1}{c|}{25.19} &
  3.12 &
  3.53 &
  \multicolumn{1}{c|}{3.87} &
  18.88 &
  20.19 &
  \multicolumn{1}{c|}{17.89} &
  21.35$\pm$.10 &
  24.74$\pm$.22 &
  27.32$\pm$.15 \\
 &
  RGB &
  RemoteCLIP  &
  28.45 &
  38.08 &
  \multicolumn{1}{c|}{37.27} &
  39.22 &
  43.05 &
  \multicolumn{1}{c|}{43.08} &
  13.19 &
  13.13 &
  \multicolumn{1}{c|}{16.73} &
  2.47 &
  4.76 &
  \multicolumn{1}{c|}{2.46} &
  18.44 &
  19.88 &
  \multicolumn{1}{c|}{17.03} &
  20.35$\pm$.14 &
  23.78$\pm$.10 &
  23.31$\pm$.11 \\
 &
  RGB &
  SkyCLIP  &
  22.41 &
  26.46 &
  \multicolumn{1}{c|}{35.13} &
  3.78 &
  4.39 &
  \multicolumn{1}{c|}{26.22} &
  2.14 &
  2.56 &
  \multicolumn{1}{c|}{3.79} &
  2.09 &
  2.11 &
  \multicolumn{1}{c|}{3.13} &
  9.14 &
  16.30 &
  \multicolumn{1}{c|}{15.13} &
  7.91$\pm$.06 &
  10.36$\pm$.10 &
  16.68$\pm$.10 \\
 &
  RGB &
  SenCLIP  &
  38.13 &
  37.86 &
  \multicolumn{1}{c|}{42.06} &
  32.04 &
  32.35 &
  \multicolumn{1}{c|}{39.23} &
  \uline{27.03} &
  23.28 &
  \multicolumn{1}{c|}{22.70} &
  3.29 &
  3.58 &
  \multicolumn{1}{c|}{4.57} &
  28.28 &
  32.50 &
  \multicolumn{1}{c|}{20.87} &
  25.75$\pm$.10 &
  25.91$\pm$.10 &
  25.89$\pm$.14 \\

   &
  MS &
  Llama3-MS-CLIP &
  45.39 &
  42.32 &
  \multicolumn{1}{c|}{26.49} &
  44.02 &
  50.67 &
  \multicolumn{1}{c|}{49.63} &
  10.79 &
  19.57 &
  \multicolumn{1}{c|}{20.98} &
  2.84 &
  2.65 &
  \multicolumn{1}{c|}{1.07} &
  18.41 &
  11.55 &
  \multicolumn{1}{c|}{33.21} &
  24.29$\pm$.12 &
  25.35$\pm$.11 &
  26.28$\pm$.13 \\ 
    &
  RGB &
  TimeSenCLIP-P64 &
  \uline{53.18} &
  \uline{53.96} &
  \multicolumn{1}{c|}{\uline{55.20}} &
  \uline{56.61} &
  \textbf{61.21} &
  \multicolumn{1}{c|}{\uline{61.01}} &
  24.16 &
  \uline{24.43} &
  \multicolumn{1}{c|}{ \textbf{31.80}} &
  \uline{7.28} &
  \uline{7.43} &
  \multicolumn{1}{c|}{\uline{13.90}} &
  \uline{32.15} &
  \uline{34.34} &
  \multicolumn{1}{c|}{\uline{33.86}} &
  \uline{34.68$\pm$.16} &
  \uline{36.27$\pm$.15} &
  \uline{39.15$\pm$.24} \\

  \hline
  
\multicolumn{21}{l}{\textbf{Single Pix}} \\ 
\hline
\multirow{2}{*}{1} &
  
  RGB &
  TimeSenCLIP-P1 &
  52.33 &
  53.59 &
  \multicolumn{1}{c|}{53.64} &
  \textbf{56.64} &
  60.31 &
  \multicolumn{1}{c|}{\uline{61.56}} &
  26.57 &
  23.84 &
  \multicolumn{1}{c|}{25.98} &
  4.76 &
  4.84 &
  \multicolumn{1}{c|}{5.36} &
  21.36 &
  22.25 &
  \multicolumn{1}{c|}{15.90} &
  32.33$\pm$.16 &
  32.97$\pm$.14 &
  32.49$\pm$.12 \\
 &
 
  MS &
  TimeSenCLIP-P1 &
  \uline{55.00} &
  \uline{56.27} &
  \multicolumn{1}{c|}{\uline{57.23}} &
  50.14 &
  \uline{59.17} &
  \multicolumn{1}{c|}{61.16} &
  \textbf{33.92} &
  \textbf{29.60} &
  \multicolumn{1}{c|}{\uline{29.42}} &
  \uline{9.17} &
  \uline{8.72} &
  \multicolumn{1}{c|}{\uline{13.39}} &
  \uline{28.08} &
  \uline{28.61} &
  \multicolumn{1}{c|}{\uline{26.09}} &
  \uline{35.26$\pm$.11} &
  \uline{36.47$\pm$.13} &
  \uline{37.46$\pm$.15} \\ \hline

\multirow{1}{*}{12} &

  MS &
  TimeSenCLIP-P1 &
  \textbf{62.40} &
  \textbf{61.35} &
  \multicolumn{1}{c|}{ \textbf{66.49}} &
  54.45 &
  60.67 &
  \multicolumn{1}{c|}{ \textbf{64.59}} &
  24.06 &
  25.01 &
  \multicolumn{1}{c|}{30.90} &
  \textbf{29.33} &
  \textbf{28.42} &
  \multicolumn{1}{c|}{ \textbf{40.36}} &
  \textbf{49.87} &
  \textbf{45.65} &
  \multicolumn{1}{c|}{\textbf{34.42}} &
  \textbf{42.02 $\pm$ .15} &
  \textbf{44.22 $\pm$.16} &
  \textbf{47.35 $\pm$ .18} \\ \bottomrule
\end{tabular}
}

\caption{Top-1 zero-shot classification accuracy (\%) for CLIP-based baselines and \textbf{TimeSenCLIP} across five geospatial tasks: Land Cover, Land Use, Habitat, Crops, and Bioregion. Results are reported using \emph{class-name (C)}, \emph{generic (G)}, and \emph{description-level (D)} textual prompts.  
All baselines operate on $64\times64$ pixel Sentinel-2 image patches. \textbf{TimeSenCLIP-P1} uses single-pixel inputs ($1\times1$) whereas \textbf{TimeSenCLIP-P64} leverages SenCLIP-pretrained spatial embeddings combined with a temporal transformer and is trained and evaluated using $64\times64$ pixel inputs. TimeSenCLIP-P1 is the only model evaluated on a 12-timestep temporal sequence; other baselines use a single timestamp. Both RGB and multispectral (MS) variants are compared.  
Values are averaged over five random seeds, with the standard deviation reported in the overall column. \uline{Underline} indicates the best performance within each pixel and temporal configuration, while \textbf{bold} denotes the best overall performance across all temporal and spatial settings.}

\label{tab:zeroshot_clip}
\end{table*}

%% file: table_ItoI_clipbased.tex
\begin{table*}[t]
\centering
\small
\setlength{\tabcolsep}{4pt}
\renewcommand{\arraystretch}{1.15}

\resizebox{\textwidth}{!}{

\begin{tabular}{lllcccccccccccccc}
\hline
\textbf{T} &
  \textbf{Bands} &
  \textbf{Model} &
  \multicolumn{2}{c|}{\textbf{Land Cover}} &
  \multicolumn{2}{c|}{\textbf{Land Use}} &
  \multicolumn{2}{c|}{\textbf{Habitat}} &
  \multicolumn{2}{c|}{\textbf{Crops}} &
  \multicolumn{2}{c|}{\textbf{Bioregion}} &
  \multicolumn{2}{c|}{\textbf{Country}} &
  \multicolumn{2}{c}{\textbf{Overall}} \\ \cline{4-17} 
 &
   &
   &
  G2S &
  \multicolumn{1}{c|}{S2G} &
  G2S &
  \multicolumn{1}{c|}{S2G} &
  G2S &
  \multicolumn{1}{c|}{S2G} &
  G2S &
  \multicolumn{1}{c|}{S2G} &
  G2S &
  \multicolumn{1}{c|}{S2G} &
  G2S &
  \multicolumn{1}{c|}{S2G} &
  G2S &
  S2G \\ \hline
\multicolumn{17}{l}{\textbf{64x64 Pix}} \\ \hline
\multirow{7}{*}{1} &
  RGB &
  CLIP  &
  0.438 &
  \multicolumn{1}{c|}{0.141} &
  0.513 &
  \multicolumn{1}{c|}{0.152} &
  0.360 &
  \multicolumn{1}{c|}{0.269} &
  0.157 &
  \multicolumn{1}{c|}{0.127} &
  0.361 &
  \multicolumn{1}{c|}{0.231} &
  0.130 &
  \multicolumn{1}{c|}{0.047} &
  0.327 &
  0.161 \\
 &
  RGB &
  GeoRSCLIP  &
  0.439 &
  \multicolumn{1}{c|}{0.313} &
  \uline{0.628} &
  \multicolumn{1}{c|}{0.274} &
  0.386 &
  \multicolumn{1}{c|}{0.241} &
  \uline{0.165} &
  \multicolumn{1}{c|}{0.087} &
  0.425 &
  \multicolumn{1}{c|}{0.263} &
  0.116 &
  \multicolumn{1}{c|}{0.077} &
  0.360 &
  0.209 \\
 &
  RGB &
  RemoteCLIP  &
  0.394 &
  \multicolumn{1}{c|}{0.406} &
  0.539 &
  \multicolumn{1}{c|}{0.496} &
  0.368 &
  \multicolumn{1}{c|}{0.188} &
  0.060 &
  \multicolumn{1}{c|}{0.078} &
  0.283 &
  \multicolumn{1}{c|}{0.200} &
  0.099 &
  \multicolumn{1}{c|}{0.025} &
  0.291 &
  0.232 \\
 &
  RGB &
  SkyCLIP  &
  0.114 &
  \multicolumn{1}{c|}{0.305} &
  0.257 &
  \multicolumn{1}{c|}{0.255} &
  0.064 &
  \multicolumn{1}{c|}{0.121} &
  0.025 &
  \multicolumn{1}{c|}{0.030} &
  0.221 &
  \multicolumn{1}{c|}{0.242} &
  0.075 &
  \multicolumn{1}{c|}{0.049} &
  0.126 &
  0.167 \\
 &
  RGB &
  SenCLIP   &
  0.483 &
  \multicolumn{1}{c|}{0.387} &
  0.589 &
  \multicolumn{1}{c|}{0.424} &
  0.342 &
  \multicolumn{1}{c|}{0.263} &
  0.103 &
  \multicolumn{1}{c|}{0.077} &
  0.473 &
  \multicolumn{1}{c|}{0.486} &
  0.183 &
  \multicolumn{1}{c|}{0.197} &
  0.362 &
  0.306 \\ 
   &
  MS &
 Llama3-MS-CLIP &
  0.267 &
  \multicolumn{1}{c|}{0.269} &
  0.301 &
  \multicolumn{1}{c|}{0.263} &
  0.317 &
  \multicolumn{1}{c|}{0.135} &
  0.050 &
  \multicolumn{1}{c|}{0.064} &
  0.276 &
  \multicolumn{1}{c|}{0.196} &
  0.098 &
  \multicolumn{1}{c|}{0.078} &
  0.218 &
  0.167 \\ 
  &
  RGB &
  TimeSenCLIP-P64 &
  \uline{0.539} &
  \multicolumn{1}{c|}{\uline{0.563}} &
  \uline{0.628} &
  \multicolumn{1}{c|}{\uline{0.635}} &
  \uline{0.404} &
  \multicolumn{1}{c|}{\uline{0.417}} &
  0.132 &
  \multicolumn{1}{c|}{\uline{0.188}} &
  \uline{0.531} &
  \multicolumn{1}{c|}{\uline{0.601}} &
  \textbf{0.261} &
  \multicolumn{1}{c|}{\uline{0.355}} &
  \uline{0.416} &
  \uline{0.460} \\ \hline

\multicolumn{17}{l}{\textbf{Single Pix}} \\ \hline
\multirow{2}{*}{1} &
  
  RGB &
  TimeSenCLIP-P1 &
  0.507 &
  \multicolumn{1}{c|}{0.534} &
  \uline{0.594} &
  \multicolumn{1}{c|}{0.601} &
  \uline{0.349} &
  \multicolumn{1}{c|}{\uline{0.340}} &
  0.125 &
  \multicolumn{1}{c|}{0.095} &
  0.295 &
  \multicolumn{1}{c|}{0.300} &
  0.114 &
  \multicolumn{1}{c|}{0.113} &
  0.331 &
  0.331 \\
 &
  MS &
  TimeSenCLIP-P1 &
  \uline{0.534} &
  \multicolumn{1}{c|}{\uline{0.574}} &
  0.573 &
  \multicolumn{1}{c|}{\uline{0.626}} &
  0.343 &
  \multicolumn{1}{c|}{0.321} &
  \uline{0.174} &
  \multicolumn{1}{c|}{\uline{0.180}} &
  \uline{0.338} &
  \multicolumn{1}{c|}{\uline{0.342}} &
  \uline{0.150} &
  \multicolumn{1}{c|}{\uline{0.146}} &
  \uline{0.352} &
  \uline{0.365}\\ \hline

\multirow{1}{*}{12} &
  
  MS &
  TimeSenCLIP-P1 &
  \textbf{0.599} &
  \multicolumn{1}{c|}{ \textbf{0.660}} &
  \textbf{0.646} &
  \multicolumn{1}{c|}{ \textbf{0.682}} &
  \textbf{0.407} &
  \multicolumn{1}{c|}{ \textbf{0.433}} &
   \textbf{0.244} &
  \multicolumn{1}{c|}{\textbf{0.411}} &
  \textbf{0.526} &
  \multicolumn{1}{c|}{ \textbf{0.634}} &
  0.255 &
  \multicolumn{1}{c|}{\textbf{0.392}} &
  \textbf{0.446} &
  \textbf{0.535} \\ \bottomrule
\end{tabular}
}

\caption{
\emph{Cross-Modal retrieval} performance across land cover, land use, habitat, crops, bioregion, and country tasks.
Results are reported as \textbf{Ground-to-Satellite (G2S)} and \textbf{Satellite-to-Ground (S2G)} Recall@1 retrieval.
Experiments are grouped by temporal sequence length ($T$), with RGB and multispectral (MS) variants compared within each temporal setting.
\uline{Underline} indicates the best performance within each pixel and temporal configuration, while \textbf{bold} denotes the best overall performance across all temporal and spatial settings.}

\label{tab:retrieval_results_clip}
\end{table*}

%% file: table_scenicness.tex
\begin{table}[t]
\centering
\footnotesize
\resizebox{\columnwidth}{!}{%

\begin{tabular}{lcccccc}
\hline
\textbf{Model} & \textbf{T} & \textbf{Pix} & \multicolumn{2}{c}{\textbf{Early}} & \multicolumn{2}{c}{\textbf{Late}} \\ \cline{4-7}

 & & & R & $\tau$ & R & $\tau$ \\
\hline
CLIP (Ground Images)\\~\cite{levering2024prompt} & 1 & -- & 0.544 & 0.390 & 0.660 & 0.465 \\
\hline
CLIP (Satellite) & 1 & 64 & 0.510 & 0.358 & 0.519 & 0.363 \\
SenCLIP & 1 & 64 & 0.396 & 0.272 & 0.421 & 0.283 \\
TimeSenCLIP-RGB & 1 & 64 & \textbf{0.532} & \textbf{0.385} & \textbf{0.660} & \textbf{0.450} \\
\hline
TimeSenCLIP-MS & 1 & 1 & 0.336 & 0.223 & 0.318 & 0.205 \\
TimeSenCLIP-MS & 4 & 1 & 0.438 & 0.309 & 0.451 & 0.312 \\
TimeSenCLIP-MS & 12 & 1 & \textbf{0.518} & \textbf{0.361} & \textbf{0.527} & \textbf{0.363} \\
\hline
\end{tabular}}

\caption{Scenicness estimation performance of TimeSenCLIP using multispectral single-pixel inputs compared to CLIP and SenCLIP. We report Early and Late ensembling, where Late columns report the average over late ensembling strategies (2, 5, 8, 10 prompts).}
\label{tab:scenicness_main}
\end{table}

%% file: table_zeroshot_baseline.tex
\begin{table*}[t]
\centering
\footnotesize
\setlength{\tabcolsep}{3pt}
\renewcommand{\arraystretch}{1.15}

\resizebox{\textwidth}{!}{
\begin{tabular}{
c |
ccc | ccc | ccc |
ccc | ccc |
ccc
}

\toprule
\multirow{2}{*}{\textbf{Model}} &

\multicolumn{3}{c|}{\textbf{Land Cover}} &
\multicolumn{3}{c|}{\textbf{Land Use}} &
\multicolumn{3}{c|}{\textbf{Habitat}} &
\multicolumn{3}{c|}{\textbf{Crops}} &
\multicolumn{3}{c|}{\textbf{Bioregion}} &
\multicolumn{3}{c}{\textbf{Overall}} \\

& 
C & G & D &
C & G & D &
C & G & D &
C & G & D &
C & G & D &
C & G & D \\

\midrule

Supervised Upper Bound & 78.65 & - & - & 79.30 & - & - & 60.52 & - & - &   64.60  & - & - & 85.39& - & - & 73.70 & - & -\\
 CNN1D              & 60.60 & 60.83 & 64.80 & 50.70 & 58.60 & 61.32 & 25.23 & 25.23 & 29.65 & 22.61 & 23.86 & 34.99 & 36.08 & 41.30 & 33.65 & 39.04$\pm$.19 & 41.96$\pm$.22 & 44.88$\pm$.15 \\
 ConvTran           & 60.29 & 60.26 & 65.06 & 52.11 & 59.91 & 62.49 & 26.59 & 25.87 & 29.81 & 23.77 & 25.07 & 38.70 & 37.55 & 42.69 & 33.38 & 40.06$\pm$.24 & 42.76$\pm$.24 & 45.89$\pm$.22 \\
 MLP                & 60.91 & 60.74 & 65.80 & 51.73 & 61.19 & 64.26 & 26.46 & 26.38 & 30.39 & 28.50 & 27.64 & 39.97 & 38.42 & 43.50 & 33.19 & 41.20$\pm$.24 & 43.89$\pm$.22 & 46.72$\pm$.14 \\
 TempCNN            & \textbf{64.30} & \textbf{63.43} & \textbf{66.53} & 53.82 & 59.30 & \textbf{65.41} & \textbf{35.70} & \textbf{33.01} & \textbf{33.37} & 23.48 & 24.66 & 34.33 & 32.42 & 36.86 & 27.41 & 41.94$\pm$.24 & 43.45$\pm$.22 & 45.41$\pm$.11 \\
 Transformer (ours) & 62.40 &
   61.35 &
  \multicolumn{1}{c|}{ 66.49} &
  \textbf{54.45} &
  \textbf{60.67} &
  \multicolumn{1}{c|}{ 64.59} &
  24.06 &
  25.01 &
  \multicolumn{1}{c|}{30.90} &
   \textbf{29.33} &
   \textbf{28.42} &
  \multicolumn{1}{c|}{ \textbf{40.36}} &
   \textbf{49.87} &
   \textbf{45.65} &
  \multicolumn{1}{c|}{ \textbf{34.42}} &
   \textbf{42.02 $\pm$ .15} &
   \textbf{44.22 $\pm$.16} &
   \textbf{47.35 $\pm$ .18} \\

\bottomrule
\end{tabular}
}
\caption{Top-1 Zero-shot accuracy across classical time-series model architecture and the proposed transformer model on tasks using Class (C), Generic (G), and Description (D)-level textual prompts. Results report the mean accuracy over five runs with different random seeds and standard deviation reported as average in overall column. Best performance is shown in \textbf{bold}.}
\label{tab:zeroshot}
\end{table*}

%% file: table_dropout_zeroshot.tex



\begin{table} 
\centering
\footnotesize
\setlength{\tabcolsep}{2pt}
\renewcommand{\arraystretch}{1.3}
\resizebox{\columnwidth}{!}{
\begin{tabular}{c c c | c c c c c c}
\toprule
\textbf{T} & \textbf{\begin{tabular}[c]{@{}c@{}}TS\\ Drop\end{tabular}} & \textbf{\begin{tabular}[c]{@{}c@{}}MS\\ Drop\end{tabular}} &
\textbf{Land Cover} & \textbf{Land Use} & \textbf{Habitat} &
\textbf{Bioregion} & \textbf{Crops} & \textbf{Average} \\
\midrule

\multirow{4}{*}{1}
 & \ding{55} & \ding{55} & 28.69 & 25.86 & 9.52 & 18.68 & 3.23 & 17.20 \\
 & \ding{55} & \ding{51}    & 23.76 & 27.88 & 8.39 & 18.18 & 2.93 & 16.23 \\
 & \ding{51}    & \ding{55} & \textbf{56.32} & \textbf{57.83} & \textbf{31.29} & \textbf{25.30} & \textbf{12.61} & \textbf{36.67} \\
 & \ding{51}    & \ding{51}    & 53.52 & 56.15 & 28.03 & 25.00 & 7.81 & 34.10 \\
\midrule


\multirow{4}{*}{12}
 & \ding{55} & \ding{55} & 61.72 & 59.76 & 23.43 & 38.71 & 31.82 & 43.09 \\
 & \ding{55} & \ding{51}    & 63.12 & 58.77 & 25.15 & 40.47 & 31.14 & 43.73 \\
 & \ding{51}    & \ding{55} & \textbf{63.63} & \textbf{60.03} & 25.54 & \textbf{40.24} & \textbf{32.23} & \textbf{44.34} \\
 & \ding{51}    & \ding{51}    & 62.42 & 58.17 & \textbf{26.68} & 39.42 & 31.64 & 43.67 \\
\bottomrule

\end{tabular}}
\caption{Effect of multispectral (MS) and temporal (TS) dropout applied during training (over 12 temporal and 10 spectral) on zero-shot classification performance (Top-1 accuracy, \%). The table reports Top-1 accuracy averaged across different textual prompt types and evaluated using both single-timestamp and monthly temporal inputs. (\ding{51}) denotes that dropout is enabled during training, whereas (\ding{55}) indicates that no dropout is applied.}
\label{tab:zeroshot_drops}
\end{table}

%% file: appendix.tex
\clearpage 
\appendix{}

\counterwithin{figure}{section}
\counterwithin{table}{section} 

\section{Appendix}

\begin{figure*}
    \includegraphics[width=\linewidth]{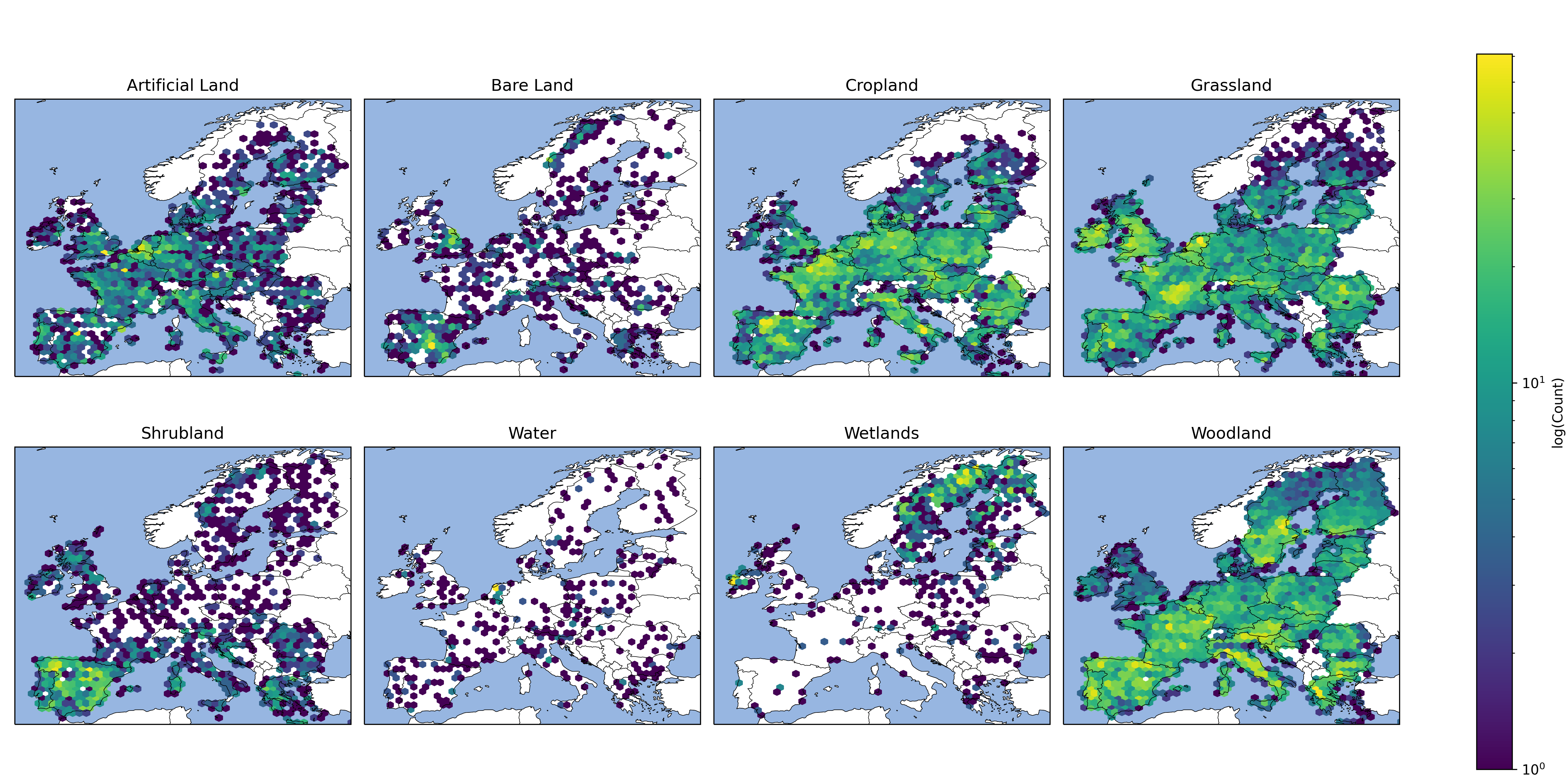}
    \caption{Land cover class distribution across  the EU in the evaluation dataset.}
    \label{fig:test_data_dist}
\end{figure*}

\begin{figure*}
        \includegraphics[width=\linewidth,scale=0.6]{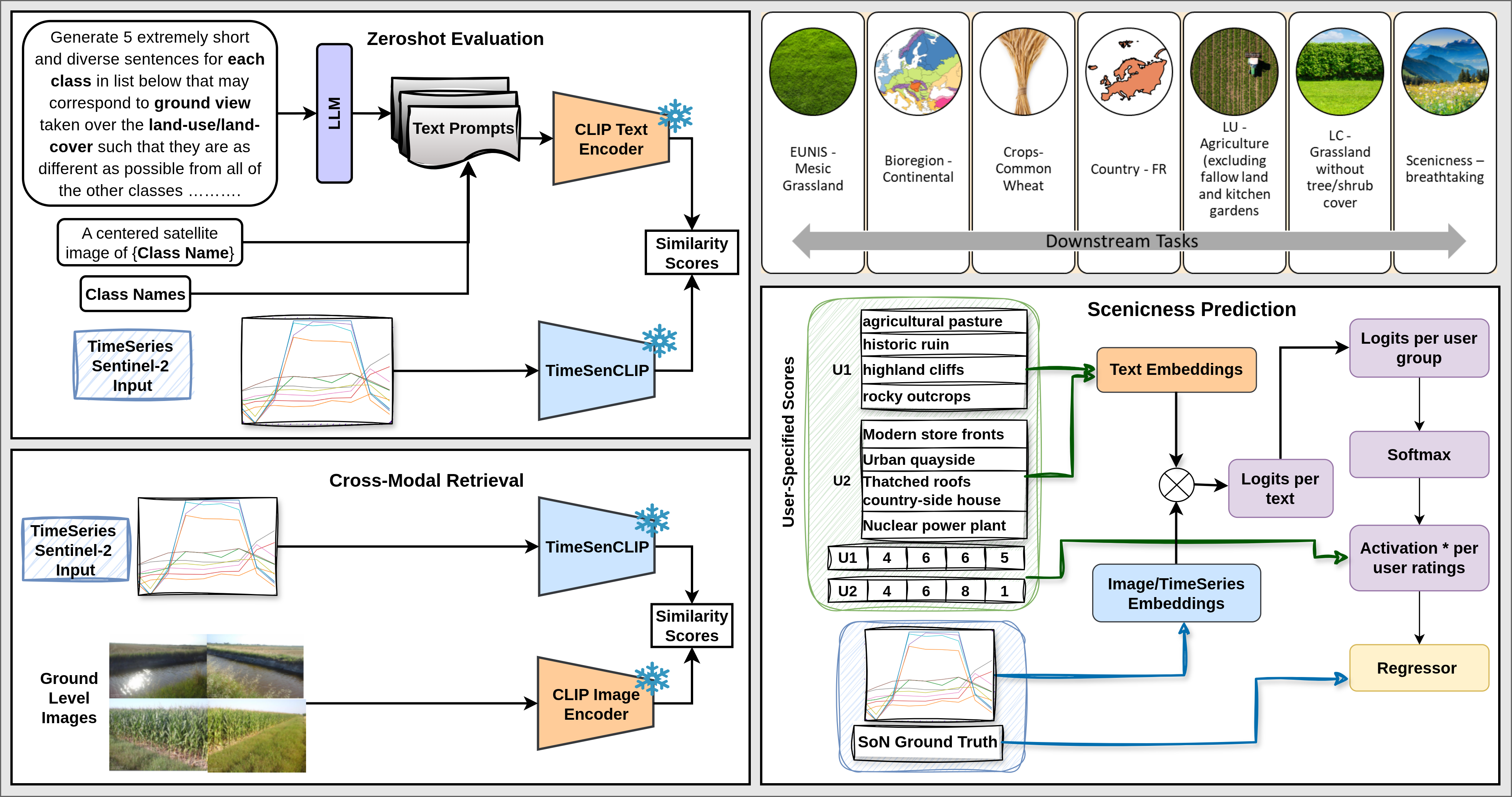}
        \caption{The infographic presents the comprehensive evaluation workflow of the TimeSenCLIP model. The pipeline integrates three key components: (1) \textbf{Zero-shot classification}, where the trained TimeSenCLIP (time series encoder) and CLIP text encoder (text prompt encoder) are used to perform inference across multiple downstream tasks. (2) \textbf{Cross-modal retrieval}, where satellite time-series and ground-level image embeddings are generated using the CLIP image encoder, and similarity scores are computed in both Satellite-to-Ground (S2G) and Ground-to-Satellite (G2S) directions to assess class-consistent retrieval performance; and (3) \textbf{Scenicness assessment}, which evaluates perceptual and aesthetic qualities of landscapes using the learned visual embeddings}
        \label{fig:evaluation_setup}
    \end{figure*}

\input{table_scenicness_full}
\input{table_ItoI_baseline}

\begin{figure}[t]
\centering
\includegraphics[width=\columnwidth]{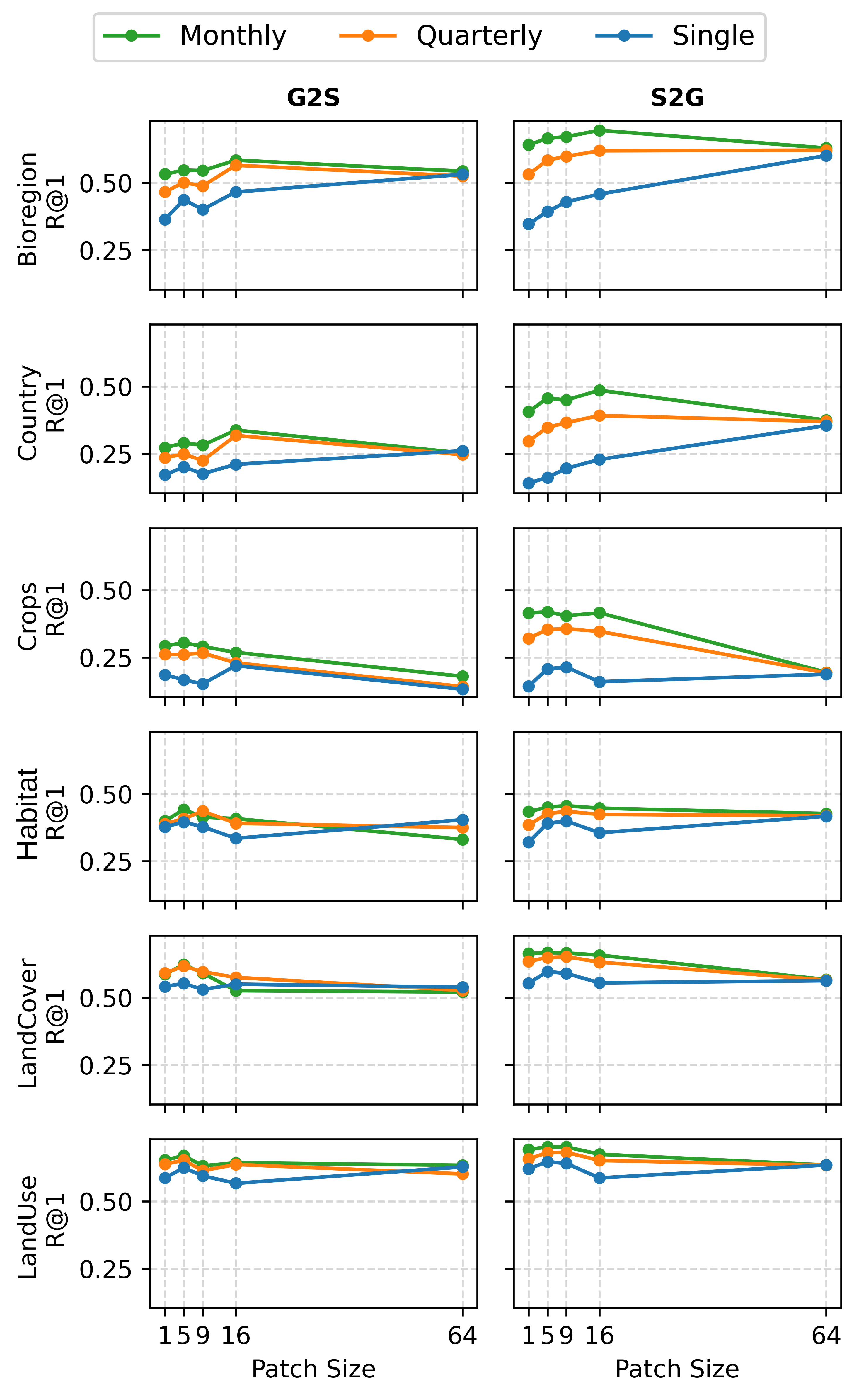}
\caption{Impact of spatial information on cross-modal retrieval for S2G and G2S tasks across different semantic categories. Trends mirror zero-shot classification, with single-pixel sequences remaining competitive.}
\label{fig:itoi_spat}
\end{figure}
\subsection{Evaluation Setup}
\label{apx:eval_setup}
The overall evaluation framework is illustrated in Figure~\ref{fig:evaluation_setup}. As shown in Section~\ref{sec:eval_setup}, we assess the proposed models across three settings that is zero-shot mapping, cross-modal retrieval, and scenicness regression. We describe the detailed evaluation protocol for each task in this section.
\subsubsection{Zero-shot Evaluation Setup}
As illustrated in Figure~\ref{fig:evaluation_setup}, zero-shot classification is performed by aligning satellite time-series data with textual prompts in a shared latent space. We utilize the frozen CLIP text encoder to transform textual prompts, ranging from simple class names and generic templates to descriptive, LLM-generated descriptions into textual embeddings. Simultaneously, the proposed TimeSenCLIP encoder processes the satellite time-series into embeddings. Both textual and time-series embeddings are $L_2$ normalized to ensure a consistent scale.The model identifies the correct class by calculating the cosine similarity between the satellite embedding and each candidate text embedding. The class corresponding to the highest similarity score is selected as the final prediction. This approach eliminates the need for task-specific labels or fine-tuning, allowing the model to generalize to novel categories based purely on their semantic descriptions.

\subsubsection{Cross-Modal Retrieval Setup}
As illustrated in Figure~\ref{fig:evaluation_setup}, cross-modal retrieval is conducted by aligning ground-level images with corresponding satellite single-pixel time-series in a joint embedding space. The ground-level images are processed through a frozen CLIP vision encoder, while the satellite time-series are encoded using the proposed TimeSenCLIP model. Both representations are $L_2$ normalized. Retrieval is performed by calculating the cosine similarity between a query embedding from one domain and a gallery of candidate embeddings from the other. For instance, in ground-to-satellite retrieval, the satellite time-series with the highest similarity to the ground-level query is selected as the top match. This capability demonstrates that TimeSenCLIP successfully captures the physical and phenological characteristics of a location, enabling accurate matching across vastly different perspectives without the need for shared metadata or geographical coordinates.
\subsubsection{Scenicness Regression Prompt Ensembling Setup}
Scenicness regression is performed using frozen TimeSenCLIP time-series embeddings together with frozen CLIP text embeddings as done with zero-shot evaluation. The evaluation follows the late prompt-ensembling strategy of \cite{levering2021relation}. In this setting (Figure~\ref{fig:evaluation_setup}), late ensembling means that the scenicness prediction is built step-by-step for each individual voter before combining the results.
First, the image (or time-series data) is converted into image embeddings, while every text description from a voter is converted into text embeddings. The model then compares the image embedding with each text embedding to produce logits per text, which measure how strongly the image matches each description.

Next, these logits are grouped per voter. For each voter, the logits are passed through a softmax function, which turns them into normalized activation values. These activations indicate how much each description contributes to that voter’s scenicness judgment. The activations are then multiplied by the user-provided rating scores, producing a voter-specific scenicness estimate for the image. Finally, the individual scenicness estimates from all voters are averaged together using a regressor to obtain the final scenicness prediction for the image.

This approach differs from early ensembling, where all text descriptions from all voters are merged into a single pool before computing logits and softmax. Early ensembling therefore produces one shared prediction immediately, while late ensembling keeps each voter’s contribution separate until the final averaging step, allowing the model to better reflect differences in human perception of scenicness.

\subsection{Scenicness Regression Analysis}
\label{apx:scennicness_full}

Table~\ref{tab:prompt_ensembling} reports the complete scenicness evaluation, including late-ensembling and early strategies, baseline CLIP variants, and TimeSenCLIP models with both single-pixel (P1-MS) and large-patch (P64-RGB) inputs. Baseline models operate on single-temporal inputs only, whereas TimeSenCLIP incorporates temporal encoding and is evaluated under single ($T{=}1$), quarterly ($T{=}4$), and monthly ($T{=}12$) temporal aggregation.

Under single-temporal input, TimeSenCLIP-P64-RGB achieves the strongest satellite-based performance ($R{=}0.634$, $\tau{=}0.437$), surpassing CLIP, GeoRSCLIP, RemoteCLIP, and SenCLIP, and approaching CLIP performance on ground-level imagery ($R{=}0.637$, $\tau{=}0.450$).
This highlights two design insights:
(1) temporal encoding substantially strengthens the SenCLIP spatial encoder by capturing seasonal and long-term landscape variation; and
(2) large spatial context remains important for perceptual attributes such as scenicness, which depend on texture, landform geometry, and spatial composition.

While CLIP remains a strong non-temporal baseline ($R{=}0.517$, $\tau{=}0.362$), it is consistently outperformed by TimeSenCLIP-P64-RGB, demonstrating the added value of temporal cues even for visually static perception tasks.
In contrast, TimeSenCLIP-P1-MS underperforms in the single-temporal regime, indicating that temporal richness alone cannot compensate for minimal spatial context when only one timestamp is available.

Increasing temporal coverage reinforces these trends.
Quarterly aggregation improves performance across configurations, and monthly aggregation yields the best overall results, with TimeSenCLIP-P64-RGB reaching $R{=}0.639$ and $\tau{=}0.440$.
Importantly, even the single-pixel TimeSenCLIP-P1-MS becomes competitive when enriched with full temporal information ($R{=}0.526$, $\tau{=}0.363$), showing that detailed temporal structure can partially offset limited spatial context.

The temporal encoding markedly enhances scenicness prediction from overhead imagery.
TimeSenCLIP-P64-RGB consistently outperforms satellite-only baselines, while single-pixel variants achieve competitive performance only when supported by dense temporal observations.
These findings underscore the central role of temporal dynamics in modeling perceptual and aesthetic landscape qualities from multispectral satellite time series.

\subsection{Conventional Temporal Baseline Comparison for Cross-Modal Retrieval}


While TimeSenCLIP's transformer architecture shows only moderate gains over classical temporal models in zero-shot classification, its advantages are more pronounced in cross-modal retrieval. As shown in Table~\ref{tab:image_retrieval_baseline}, TimeSenCLIP generally outperforms CNN1D, ConvTran, MLP, and TempCNN for Satellite-to-Ground (S2G) retrieval, effectively capturing temporal patterns that aid cross-view alignment. For Ground-to-Satellite (G2S) retrieval, some conventional architectures, such as CNN1D or MLP, perform similarly or slightly better for finer-grained tasks like Land Cover, Land Use, Habitat, and Crops, likely because local ground images are easier to match to the broader satellite context. In contrast, for coarser categories such as Bioregion and Country, the transformer’s ability to integrate long-term temporal sequences provides a stable signal, improving G2S performance and highlighting the advantages of temporal modeling for large-scale, temporally coherent classes. In general, TimeSenCLIP often reverses the typical S2G/G2S performance gap, outperforming baselines in S2G retrieval and occasionally exceeding G2S accuracy, suggesting its learned embeddings form a stable semantic anchor that supports accurate cross-view localization even under drastic viewpoint differences.

Intuitively, TimeSenCLIP bridges the gap between local ground observations and satellite imagery: it encodes distinctive spectral-temporal patterns while aligning them with rich semantic signals from geo-tagged ground images. This combination enables robust retrieval across diverse land-cover, land-use, ecological, and crop categories, without relying on large spatial patches or extra paired image-text supervision. Its transformer backbone and cross-view training strategy produce embeddings that are both semantically meaningful and generalizable, effectively balancing spatial and temporal cues according to task-specific requirements.

\subsection{Ablation}
\subsubsection{Impact of Spatial Context}
We report the impact of spatial context for cross-modal retrieval in Figure~\ref{fig:itoi_spat}. Results in both Ground-to-Satellite (G2S) and Satellite-to-Ground (S2G) directions broadly mirror the behavior observed in zero-shot classification, indicating that the balance between spatial and temporal context transfers consistently across evaluation settings.

Single-pixel (P1) time series already achieve strong retrieval accuracy across most tasks, highlighting the discriminative strength of temporal signatures for cross-view alignment. Incorporating limited spatial context through small patches (e.g., $5\times5$ to $16\times16$) yields task-dependent improvements, particularly for categories with structured spatial organization such as land cover and land use. Expanding the spatial extent beyond this range provides minimal additional benefit and is mainly advantageous for coarse or spatially homogeneous classes, where broader context improves stability rather than fine-grained discrimination.

Unlike zero-shot evaluation, which remains largely symmetric, retrieval reveals mild differences between G2S and S2G. G2S performance is relatively insensitive to patch size, with temporal depth accounting for most gains and larger patches helping only in coarse categories (e.g., Bioregion or Country). In contrast, S2G retrieval shows greater sensitivity to spatial context: small patches can improve alignment when ground queries contain meaningful structure, whereas overly large patches may introduce irrelevant variation and slightly reduce accuracy. Nevertheless, temporally rich P1 sequences remain competitive across all tasks and directions.

The variation across patch sizes is modest compared with the influence of temporal coverage, confirming temporal dynamics as the primary factor governing cross-view retrieval. In several settings, sufficiently long P1 time series approach the performance of substantially larger spatial contexts, demonstrating that strong temporal information can compensate for limited spatial extent and enabling scalable retrieval without large image patches.

\subsubsection{Impact of Dropout Strategies}
\input{table_dropout_retrieval}

For cross-modal retrieval, we extend the analysis to include both single, quarterly, and monthly temporal aggregations. Retrieval performance, averaged across Ground-to-Satellite (G2S) and Satellite-to-Ground (S2G) directions, follows the same trends observed in zero-shot classification (Table~\ref{tab:dropout_effect_retrieval}). Temporal dropout consistently improves alignment when temporal coverage is limited, with Recall@1 increasing from 0.219 to 0.369 for single timestamps. Incorporating MS Drop does not yield complementary benefits and can even harm performance in tasks reliant on spectral information, such as Crops and Habitat.

While TS Drop provides the largest gains for single timestamps, its effect diminishes as temporal depth increases to quarterly or monthly sequences. Single-pixel multispectral time series still achieve competitive retrieval performance, demonstrating that temporal information is the dominant factor for cross-view alignment, while spatial and spectral cues provide additional task-dependent gains.

\subsubsection{Impact of Ground Image Aggregation}

We evaluate two strategies for incorporating multiple ground-level images per location: Average Pooling, which aggregates embeddings across all available images, and Single Image, which uses one randomly selected image without pooling. As summarized in Table~\ref{tab:pooling}, Average Pooling consistently performs slightly better than the Single Image setting across most tasks. However, the overall performance gap is negligible, with the mean Top-1 accuracy varying only marginally from 41.24 to 41.30.

These findings suggest that, although aggregating multiple ground images can provide minor improvements, a single representative image is sufficient to achieve nearly identical retrieval performance. This observation reduces the reliance on collecting and processing multiple images per location, thereby simplifying data requirements and computational overhead. Importantly, it supports a more scalable training and deployment protocol for large-scale cross-view remote sensing applications, where dense ground-image coverage is often impractical.

\begin{table}[tp]
\setlength{\tabcolsep}{3.5pt}
\renewcommand{\arraystretch}{1.15}
\resizebox{\columnwidth}{!}{
\begin{tabular}{l|l|l|l|l|l|l}
\hline
{Pooling} &
  \multicolumn{1}{c|}{\begin{tabular}[c]{@{}c@{}}Land \\ Cover\end{tabular}} &
  \multicolumn{1}{c|}{\begin{tabular}[c]{@{}c@{}}Land \\ Use\end{tabular}} &
  \multicolumn{1}{c|}{Habitat} &
  \multicolumn{1}{c|}{Crops} &
  \multicolumn{1}{c|}{Bioregion} &
  \multicolumn{1}{c}{Average} \\ \hline
Average   & 60.85      & 58.65    & 29.02   & 23.53 & 34.46     & 41.30   \\
Single image        & 60.07      & 57.25    & 30.23   & 24.47 & 34.17     & 41.24   \\ \hline
\end{tabular}}

\caption{Comparison of ground image aggregation strategies. “Single image” uses a single randomly selected image from four available images, while “Average Pooling” computes the mean embedding across all four images.}
\label{tab:pooling}
\end{table}
\subsubsection{Zero-shot Class-wise Classification}
\input{cm_fig}
To provide further insight, we include a class-wise zero-shot classification behavior. Figure~\ref{fig:cm_plots} shows confusion matrices for four zero-shot classification tasks: Land Cover, Land Use, Crops, and Habitat, using TimeSenCLIP-P1-MS (monthly) with descriptive prompts. For readability, we display the top classes per task (8-Land Use and 10-for Crops and Habitat) and group remaining categories as “Other”. Overall, dominant and well-defined classes are predicted reliably, while most errors occur between semantically or structurally similar categories, reflecting intrinsic ambiguity rather than systematic failure. For example, in Crops, confusion is concentrated among phenologically similar cereals such as wheat and barley, whereas in Land Cover, shrubland is occasionally misclassified as woodland due to overlapping seasonal signatures. Habitat-level errors primarily arise between ecologically related classes, such as grasslands and agricultural areas, which share similar vegetation structure and temporal patterns.

%% file: table_scenicness_full.tex
\begin{table*}[t]
\centering
\footnotesize

\resizebox{\textwidth}{!}{%
\begin{tabular}{llcccccccccccc}
\hline
\textbf{T} &
  \textbf{Model} &
  \multicolumn{2}{c|}{\textbf{Early}} &
  \multicolumn{2}{c|}{\textbf{2}} &
  \multicolumn{2}{c|}{\textbf{5}} &
  \multicolumn{2}{c|}{\textbf{8}} &
  \multicolumn{2}{c|}{\textbf{10}} &
  \multicolumn{2}{c}{\textbf{Avg}} \\
 &
   &
   
  R &
  \multicolumn{1}{c|}{$\tau$} &
  R &
  \multicolumn{1}{c|}{$\tau$} &
  R &
  \multicolumn{1}{c|}{$\tau$} &
  R &
  \multicolumn{1}{c|}{$\tau$} &
  R &
  \multicolumn{1}{c|}{$\tau$} &
  R &
  $\tau$ \\ \hline
\multicolumn{14}{l}{\textbf{Baselines (RGB and 64 Pix)}} \\ \hline
1 &
  \begin{tabular}[c]{@{}l@{}}CLIP (SON Images) \\ \cite{levering2024prompt}\end{tabular} &
  
  0.544 &
  \multicolumn{1}{c|}{0.390} &
  0.692 &
  \multicolumn{1}{c|}{0.485} &
  0.654 &
  \multicolumn{1}{c|}{0.447} &
  0.673 &
  \multicolumn{1}{c|}{0.479} &
  0.623 &
  \multicolumn{1}{c|}{0.450} &
  0.637 &
  0.450 \\ \hline
\multirow{5}{*}{1} &
  CLIP &
  0.510 &
  \multicolumn{1}{c|}{0.358} &
  0.547 &
  \multicolumn{1}{c|}{0.384} &
  0.547 &
  \multicolumn{1}{c|}{0.384} &
  0.536 &
  \multicolumn{1}{c|}{0.379} &
  0.444 &
  \multicolumn{1}{c|}{0.307} &
  0.517 &
  0.362 \\
 &
  GeoRSCLIP &
  0.478 &
  \multicolumn{1}{c|}{0.346} &
  0.520 &
  \multicolumn{1}{c|}{0.378} &
  0.512 &
  \multicolumn{1}{c|}{0.369} &
  0.466 &
  \multicolumn{1}{c|}{0.336} &
  0.306 &
  \multicolumn{1}{c|}{0.212} &
  0.456 &
  0.328 \\
 &
  RemoteCLIP &
  0.435 &
  \multicolumn{1}{c|}{0.295} &
  0.477 &
  \multicolumn{1}{c|}{0.321} &
  0.473 &
  \multicolumn{1}{c|}{0.318} &
  0.478 &
  \multicolumn{1}{c|}{0.325} &
  0.397 &
  \multicolumn{1}{c|}{0.260} &
  0.452 &
  0.304 \\
 &
  SkyCLIP &
  0.365 &
  \multicolumn{1}{c|}{0.249} &
  0.207 &
  \multicolumn{1}{c|}{0.124} &
  0.246 &
  \multicolumn{1}{c|}{0.154} &
  0.245 &
  \multicolumn{1}{c|}{0.153} &
  0.391 &
  \multicolumn{1}{c|}{0.255} &
  0.291 &
  0.187 \\
 &
  SenCLIP &
  0.396 &
  \multicolumn{1}{c|}{0.272} &
  0.459 &
  \multicolumn{1}{c|}{0.310} &
  0.440 &
  \multicolumn{1}{c|}{0.297} &
  0.420 &
  \multicolumn{1}{c|}{0.286} &
  0.366 &
  \multicolumn{1}{c|}{0.241} &
  0.416 &
  0.281 \\ \hline
\multicolumn{14}{l}{\textbf{TimeSenCLIP (ours)}} \\ \hline

\multirow{2}{*}{1} &
  P1-MS &
  0.336 &
  \multicolumn{1}{c|}{0.223} &
  0.345 &
  \multicolumn{1}{c|}{0.231} &
  0.346 &
  \multicolumn{1}{c|}{0.231} &
  0.308 &
  \multicolumn{1}{c|}{0.200} &
  0.274 &
  \multicolumn{1}{c|}{0.167} &
  0.322 &
  0.210 \\
 &
  P64-RGB &
  \uline{ \textbf{0.532}} &
  \multicolumn{1}{c|}{\uline{ \textbf{0.385}}} &
  \textbf{0.681} &
  \multicolumn{1}{c|}{\textbf{0.461}} &
  \textbf{0.672} &
  \multicolumn{1}{c|}{\textbf{0.460}} &
  \textbf{0.671} &
  \multicolumn{1}{c|}{\textbf{0.461}} &
  \textbf{0.614} &
  \multicolumn{1}{c|}{\textbf{0.420}} &
  \textbf{0.634} &
  \textbf{0.437} \\ \hline
\multirow{2}{*}{4} &
  P1-MS &
  0.438 &
  \multicolumn{1}{c|}{0.309} &
  0.479 &
  \multicolumn{1}{c|}{0.324} &
  0.468 &
  \multicolumn{1}{c|}{0.325} &
  0.445 &
  \multicolumn{1}{c|}{0.313} &
  0.413 &
  \multicolumn{1}{c|}{0.288} &
  0.449 &
  0.312 \\
 &
  P64-RGB &
  \textbf{0.514} &
  \multicolumn{1}{c|}{\textbf{0.374}} &
  \textbf{0.682} &
  \multicolumn{1}{c|}{\textbf{0.461}} &
  \textbf{0.672} &
  \multicolumn{1}{c|}{\textbf{0.460}} &
  \textbf{0.671} &
  \multicolumn{1}{c|}{\textbf{0.460}} &
  \textbf{0.617} &
  \multicolumn{1}{c|}{\textbf{0.420}} &
  \textbf{0.631} &
  \textbf{0.435} \\ \hline
\multirow{2}{*}{12} &
  P1-MS &
  0.518 &
  \multicolumn{1}{c|}{0.361} &
  0.558 &
  \multicolumn{1}{c|}{0.371} &
  0.548 &
  \multicolumn{1}{c|}{0.373} &
  0.526 &
  \multicolumn{1}{c|}{0.366} &
  0.478 &
  \multicolumn{1}{c|}{0.343} &
  0.526 &
  0.363 \\
&
  P64-RGB &
  0.518 &
  \multicolumn{1}{c|}{\textbf{0.378}} &
  \uline{ \textbf{0.690}} &
  \multicolumn{1}{c|}{\uline{ \textbf{0.466}}} &
  \uline{ \textbf{0.681}} &
  \multicolumn{1}{c|}{\uline{ \textbf{0.465}}} &
  \uline{ \textbf{0.681}} &
  \multicolumn{1}{c|}{\uline{ \textbf{0.466}}} &
  \uline{ \textbf{0.624}} &
  \multicolumn{1}{c|}{\uline{ \textbf{0.423}}} &
  \uline{ \textbf{0.639}} &
  \uline{ \textbf{0.440}} \\ \hline
\end{tabular}
}
\caption{Scenicness estimation on UK data points using Sentinel-2 satellite imagery. Performance is reported in terms of Pearson's R and Kendall’s Tau ($\tau$) across different temporal resolutions: Single, Quarterly (4 months), and Monthly (12 months). Multiple columns correspond to different prompt ensembling strategies as proposed by Levering et al.2024: \textbf{Early}, 2, 5, 8, and 10, followed by the average (\textbf{Avg}) across all strategies. The first row shows CLIP performance on ground-level images (SoN dataset) for reference. \uline{\textbf{Bold-underline}} text indicates the best overall performance across all temporal settings, while \textbf{bold} text indicates the best performance within each temporal resolution. This table demonstrates that TimeSenCLIP, using either multispectral single-pixel (P1-MS) or larger RGB patches (P64-RGB), can achieve competitive scenicness prediction performance compared to models trained on ground-level images.}
\label{tab:prompt_ensembling}
\end{table*}

%% file: table_ItoI_baseline.tex
\begin{table*}[t]
\centering
\footnotesize
\setlength{\tabcolsep}{4pt}
\renewcommand{\arraystretch}{1.15}

\resizebox{\textwidth}{!}{
\begin{tabular}{c |
cc|cc|cc|cc|cc|cc|cc}
\hline
\textbf{Model} &
\multicolumn{2}{c|}{\textbf{Land Cover}} &
\multicolumn{2}{c|}{\textbf{Land Use}} &
\multicolumn{2}{c|}{\textbf{Habitat}} &
\multicolumn{2}{c|}{\textbf{Crops}} &
\multicolumn{2}{c|}{\textbf{Bioregion}} &
\multicolumn{2}{c|}{\textbf{Country}} &
\multicolumn{2}{c}{\textbf{Overall}} \\

 & 
G2S & S2G & 
G2S & S2G &
G2S & S2G &
G2S & S2G &
G2S & S2G &
G2S & S2G &
G2S & S2G \\
\hline

CNN1D 
& \textbf{0.632} & 0.506 
& 0.656 & 0.492 
& 0.407 & 0.303 
& 0.283 & 0.128 
& 0.366 & 0.442 
& 0.202 & 0.181 
& 0.424 & 0.342 \\

ConvTran 
& 0.618 & 0.492 
& \textbf{0.684} & 0.484 
& 0.370 & 0.266 
& \textbf{0.297} & 0.117 
& 0.372 & 0.410 
& 0.228 & 0.143 
& 0.428 & 0.319 \\

MLP 
& 0.568 & 0.583 
& 0.674 & 0.610 
& \textbf{0.413} & 0.355 
& 0.295 & 0.207 
& 0.421 & 0.510 
& 0.210 & 0.250 
& 0.430 & 0.419 \\

TempCNN 
& 0.542 & 0.573 
& 0.657 & 0.594 
& 0.394 & 0.286 
& 0.263 & 0.167 
& 0.403 & 0.475 
& 0.192 & 0.250 
& 0.409 & 0.391 \\

Transformer (ours) 
& 0.598 & \textbf{0.666} 
& 0.642 & \textbf{0.695} 
& 0.378 & \textbf{0.432} 
& 0.283 & \textbf{0.434} 
& \textbf{0.534} & \textbf{0.648} 
& \textbf{0.253} & \textbf{0.416} 
& \textbf{0.448} & \textbf{0.549} \\

\hline
\end{tabular}}
\caption{Cross-Modal retrieval performance for Satellite-to-Ground (S2G) and Ground-to-Satellite (G2S) retrieval across temporal model architectures, reported using Recall@1. \textbf{Bold} values indicate the best-performing model for each task and retrieval direction.}
\label{tab:image_retrieval_baseline}
\end{table*}

%% file: table_dropout_retrieval.tex


\begin{table}[tp]
\label{tab:sequence_ablation}
\footnotesize
\setlength{\tabcolsep}{2pt}
\renewcommand{\arraystretch}{1.3}
\resizebox{\columnwidth}{!}{
\begin{tabular}{lcc|ccccccl}
\toprule
\textbf{T} & \textbf{\begin{tabular}[c]{@{}c@{}}TS\\ Drop\end{tabular}} & \textbf{\begin{tabular}[c]{@{}c@{}}MS\\ Drop\end{tabular}}  & \textbf{Land Cover} & \textbf{Land Use} & \textbf{Habitat} & \textbf{Bioregion} & \textbf{Crops} & \textbf{Country} & \textbf{Average} \\
\midrule
\multirow{4}{*}{1} 
& \ding{55} & \ding{55} & 0.290 & 0.402 & 0.197 & 0.243 & 0.090 & 0.096 & 0.219 \\
& \ding{55} & \ding{51} & 0.210 & 0.358 & 0.129 & 0.248 & 0.114 & 0.119 & 0.196 \\
& \ding{51} & \ding{55} & 0.548 & \textbf{0.609} & \textbf{0.343} & \textbf{0.353} & \textbf{0.203} & \textbf{0.159} & \textbf{0.369} \\
& \ding{51} & \ding{51} & \textbf{0.556} & 0.608 & 0.335 & 0.344 & 0.169 & 0.152 & 0.360 \\
\midrule

\multirow{4}{*}{4} 
& \ding{55} & \ding{55} & 0.606 & 0.650 & 0.392 & 0.501 & 0.298 & \textbf{0.268} & 0.452 \\
& \ding{55} & \ding{51} & 0.607 & 0.650 & 0.390 & 0.467 & 0.262 & 0.239 & 0.436 \\
& \ding{51} & \ding{55} & \textbf{0.628} & 0.654 & 0.384 & 0.505 & \textbf{0.308} & 0.267 & \textbf{0.457} \\
& \ding{51} & \ding{51} & 0.627 & \textbf{0.658} & \textbf{0.393} & \textbf{0.508} & 0.272 & 0.261 & 0.453 \\
\midrule

\multirow{4}{*}{12} 
& \ding{55} & \ding{55} & \textbf{0.642} & \textbf{0.674} & \textbf{0.417} & 0.605 & \textbf{0.372} & \textbf{0.389} & \textbf{0.516} \\
& \ding{55} & \ding{51} & 0.623 & 0.667 & 0.413 & \textbf{0.610} & 0.356 & 0.365 & 0.505 \\
& \ding{51} & \ding{55} & 0.632 & 0.669 & 0.405 & 0.591 & 0.359 & 0.335 & 0.498 \\
& \ding{51} & \ding{51} & 0.631 & 0.665 & 0.409 & 0.584 & 0.337 & 0.327 & 0.492 \\
\bottomrule
\end{tabular}}
\caption{Effect of Multispectral (MS) and Temporal (TS) Dropout on retrieval performance (Recall@1). The table reports Recall averaged across S2G and G2S. (\ding{51}) indicates dropout applied; (\ding{55}) indicates no dropout.}
\label{tab:dropout_effect_retrieval}
\end{table}

%% file: cm_fig.tex
\begin{figure*}[t]
    \centering
    \begin{tabular}{cc}  
        \includegraphics[width=0.47\linewidth]{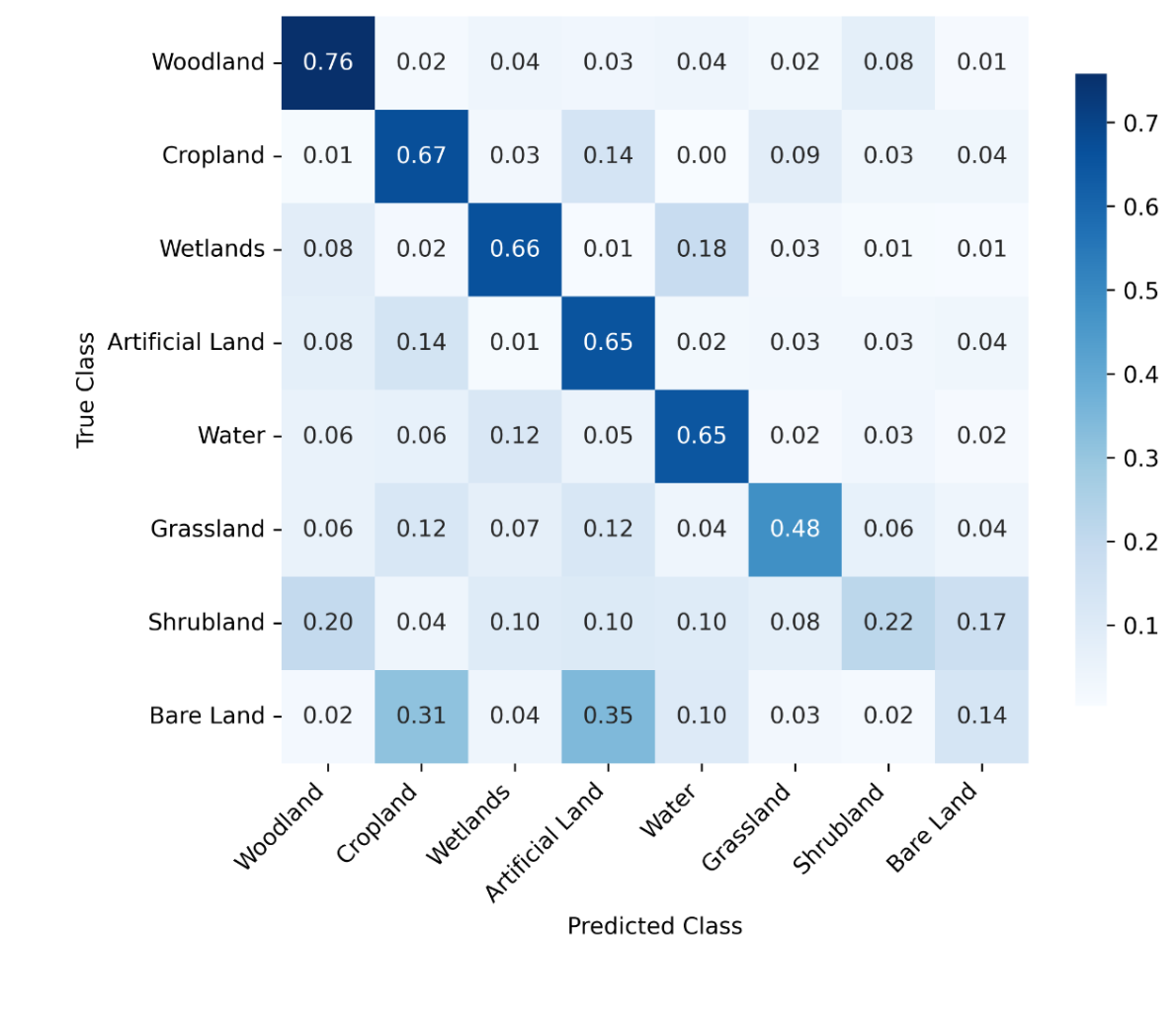} &
        \includegraphics[width=0.49\linewidth]{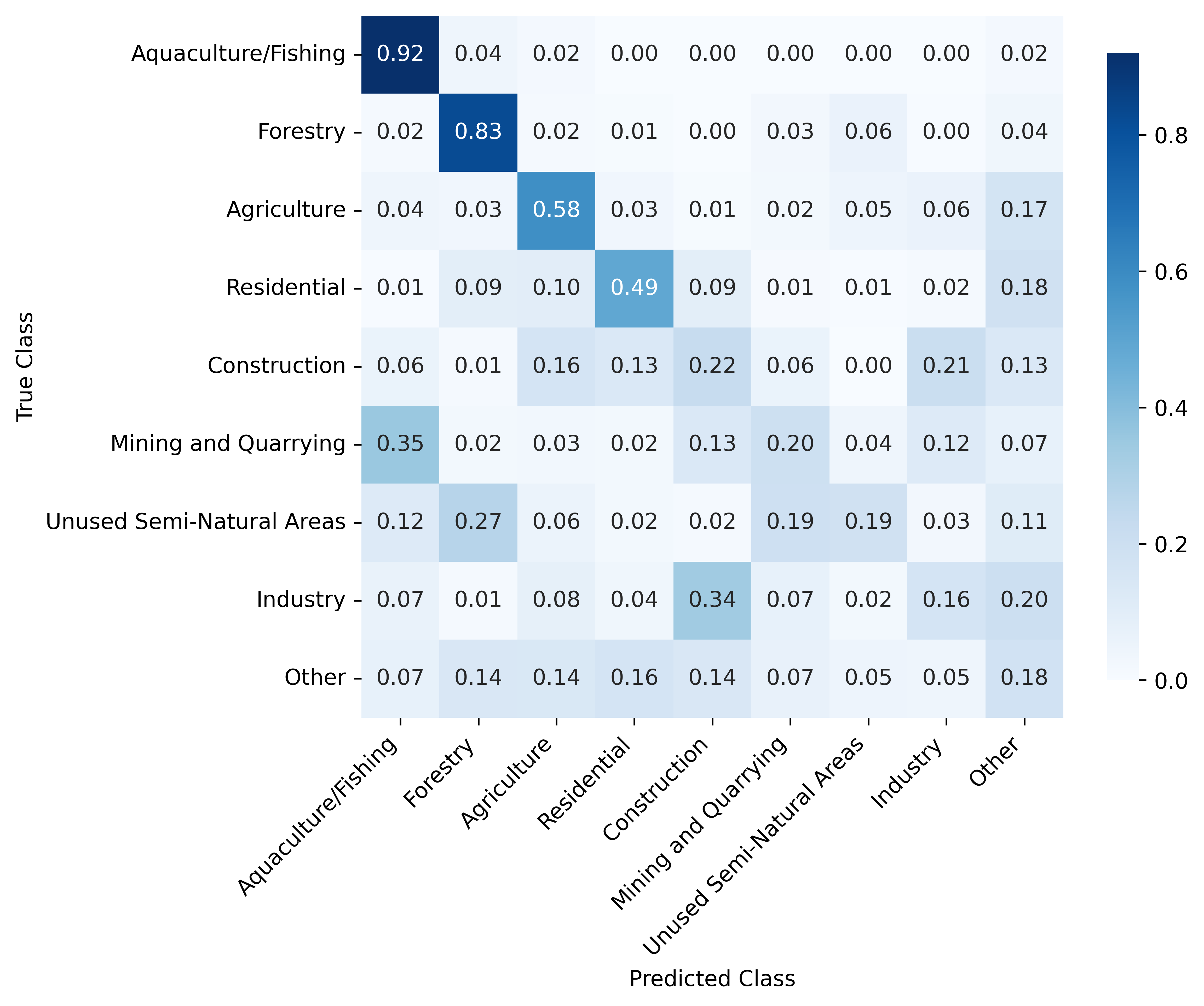} \\
        (a) Land Cover & (b) Land Use \\[3ex]
        \includegraphics[width=0.48\linewidth]{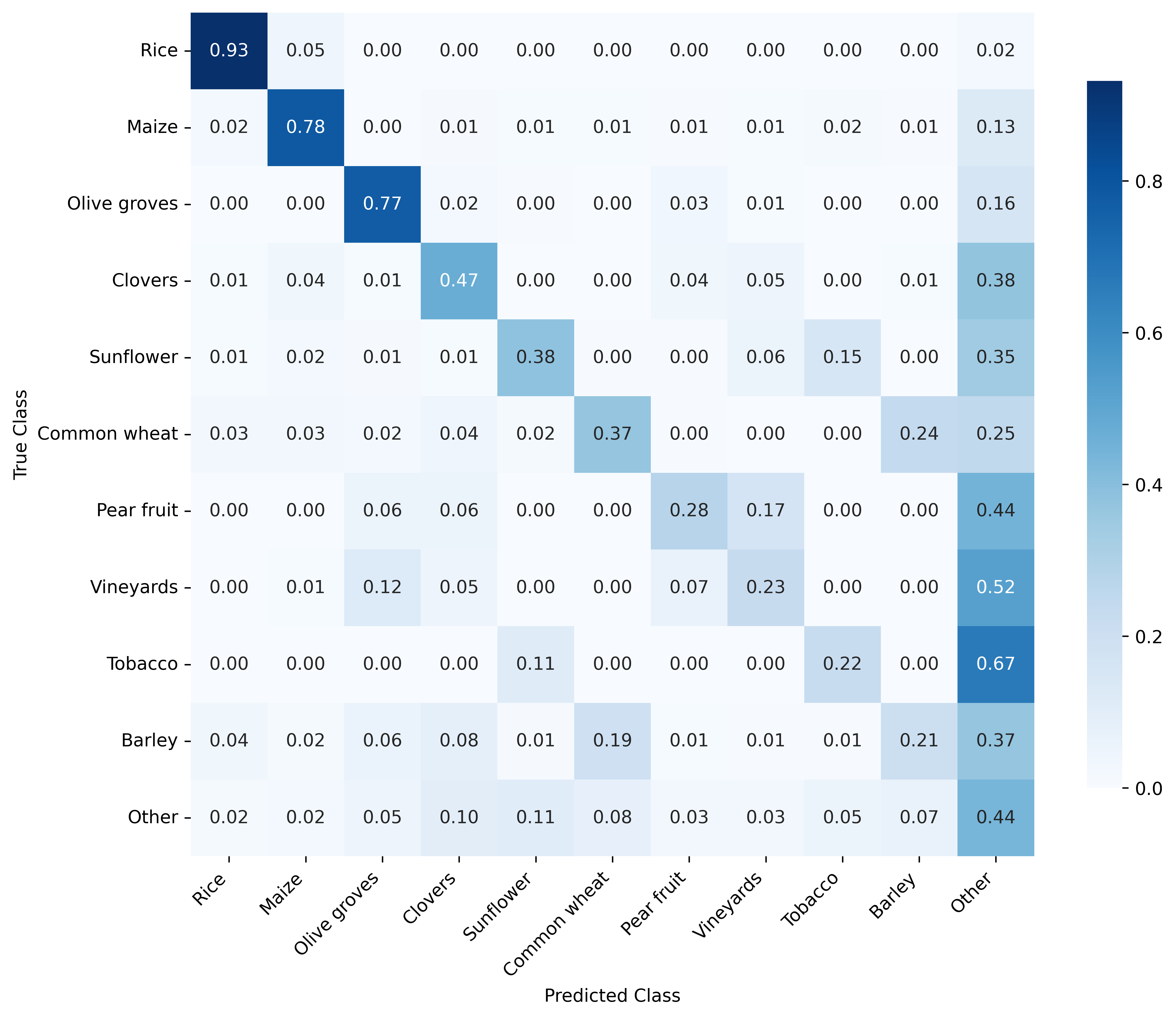} &
        \includegraphics[width=0.48\linewidth]{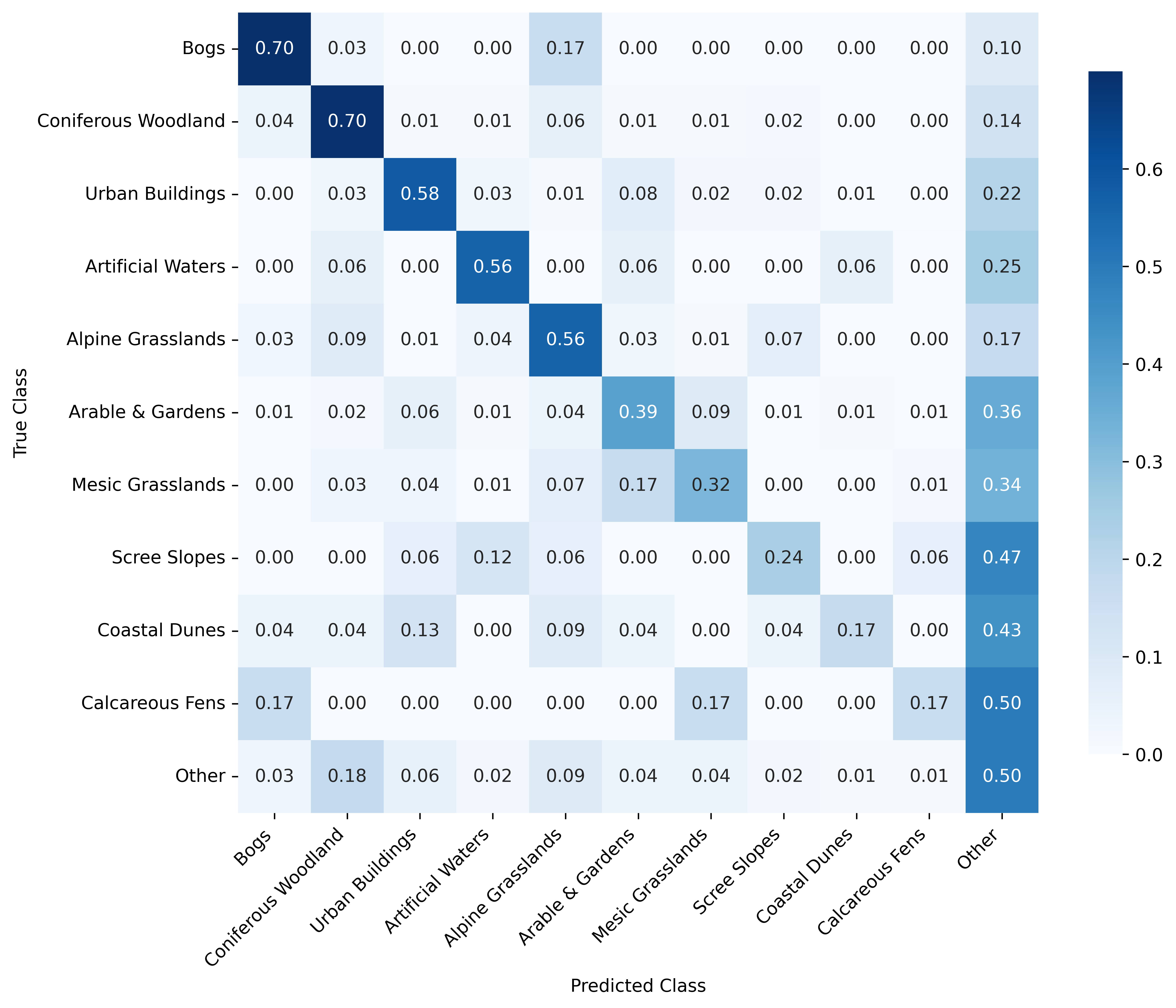} \\
        (c) Crops & (d) Habitat
    \end{tabular}
    \caption{Normalized confusion matrices for TimeSenCLIP, showing the top 8 predicted classes for LULC and the top 10 predicted classes for Crops and Habitat. Classes beyond these top predictions are grouped as ``Other”. All evaluations are performed in a zeroshot setting on monthly, single-pixel TimeSenCLIP inputs using descriptive text prompts.}
    \label{fig:cm_plots}
\end{figure*}